\documentclass[sigconf]{acmart}

\usepackage{algorithm}
\usepackage{algpseudocode}
\usepackage{bbm}
\usepackage{xcolor}
\usepackage{subcaption}
\DeclareMathOperator*{\argmax}{arg\,max}

\DeclareMathOperator{\Tr}{Tr}

\AtBeginDocument{%
  }

\setcopyright{acmcopyright}
\copyrightyear{2023}
\acmYear{2023}
\acmDOI{XXXXXXX.XXXXXXX}

\acmConference[Conference acronym 'XX]{Make sure to enter the correct
  conference title from your rights confirmation email}{June 03--05,
  2018}{Woodstock, NY}
\acmPrice{15.00}
\acmISBN{978-1-4503-XXXX-X/18/06}

\acmSubmissionID{123-A56-BU3}

\begin{document}

%%
%% The "title" command has an optional parameter,
%% allowing the author to define a "short title" to be used in page headers.
\title{An End-to-End Framework for Marketing Effectiveness Optimization under Budget Constraint}

%%
%% The "author" command and its associated commands are used to define
%% the authors and their affiliations.
%% Of note is the shared affiliation of the first two authors, and the
%% "authornote" and "authornotemark" commands
%% used to denote shared contribution to the research.
\author{Ziang Yan, Shusen Wang, Guorui Zhou, Jingjian Lin, Peng Jiang}
% \authornote{First two authors contribute equally to this research.}
\affiliation{%
  \institution{Kuaishou Inc}
  \state{Beijing}
  \country{China}
}
\email{{yanziang,wangshusen03,zhouguorui,linjingjian,jiangpeng}@kuaishou.com}
%%
%% By default, the full list of authors will be used in the page
%% headers. Often, this list is too long, and will overlap
%% other information printed in the page headers. This command allows
%% the author to define a more concise list
%% of authors' names for this purpose.
\renewcommand{\shortauthors}{Ziang Yan et al.}

%%
%% The abstract is a short summary of the work to be presented in the
%% article.
\begin{abstract}
Online platforms often incentivize consumers to improve user engagement and platform revenue.
Since different consumers might respond differently to incentives, individual-level budget allocation is an essential task in marketing campaigns.
Recent advances in this field often address the budget allocation problem using a two-stage paradigm: the first stage estimates the individual-level treatment effects using causal inference algorithms, and the second stage invokes integer programming techniques to find the optimal budget allocation solution.
Since the objectives of these two stages might not be perfectly aligned, such a two-stage paradigm could hurt the overall marketing effectiveness.

In this paper, we propose a novel end-to-end framework to directly optimize the business goal under budget constraints.
Our core idea is to construct a regularizer to represent the marketing goal and optimize it efficiently using gradient estimation techniques.
As such, the obtained models can learn to maximize the marketing goal directly and precisely.
We extensively evaluate our proposed method in both offline and online experiments, and experimental results demonstrate that our method outperforms current state-of-the-art methods.
Our proposed method is currently deployed to allocate marketing budgets for hundreds of millions of users on a short video platform and achieves significant business goal improvements.
Our code will be publicly available.
\end{abstract}

%%
%% The code below is generated by the tool at http://dl.acm.org/ccs.cfm.
%% Please copy and paste the code instead of the example below.
%%
  

%%
%% Keywords. The author(s) should pick words that accurately describe
%% the work being presented. Separate the keywords with commas.

% \received{20 February 2007}
% \received[revised]{12 March 2009}
% \received[accepted]{5 June 2009}

%%
%% This command processes the author and affiliation and title
%% information and builds the first part of the formatted document.
\maketitle

\section{Introduction}\label{sec:introduction}
Offering incentives (e.g., cash bonuses, discounts, coupons) to consumers are effective ways for online platforms to acquire new users, increase user engagement, and boost platform revenue~\cite{lin2017monetary,zhao2019uplift,zhao2019unified,makhijani2019lore,goldenberg2020free,zhang2021bcorle,tu2021personalized,ai2022lbcf,albert2022commerce,zhou2023direct}.
For example, coupons in Taobao~\cite{zhang2021bcorle} are offered to increase user activity, promotions in Booking~\cite{goldenberg2020free} can improve user satisfaction, cash bonuses in Kuaishou~\cite{ai2022lbcf} are used to stimulate user retention, and promotions in Uber~\cite{zhao2019uplift} can encourage users to start using a new product.
Despite the effectiveness, these marketing campaigns could incur high costs, and thus the total budget is often capped in real-world scenarios.
From the perspective of online platforms, the goal of a marketing campaign is often described as maximizing a specific business goal (e.g., user retention) under a specific budget constraint.
To this end, allocating an appropriate amount of incentive to different users is essential for optimizing marketing effectiveness since different users could respond differently to incentives.
This budget allocation problem has great practical implications and has been investigated for decades.

Many recent advances address the budget allocation problem using a two-stage paradigm~\cite{zhao2019unified,ai2022lbcf,tu2021personalized,albert2022commerce}: in the first stage the individual-level heterogeneous treatment effects are estimated using causal inference techniques, and then the estimated effects are fed into an integer programming formulation as coefficients to find the optimal allocation.
However, the objectives of these two stages are fundamentally different, and in practice, we have observed these objectives are not perfectly aligned.
For example, if the total budget is extremely low, then an effective allocation algorithm should distribute incentives only to a small subset of users who are fairly sensitive to incentives.
In this case, an improved treatment effects predictor might have worse marketing effectiveness, if the model has a higher precision on all users on average but performs poorly on this small subset of users.

In this paper, we propose a novel end-to-end approach to optimize marketing effectiveness under budget constraints, directly and precisely.
Our core idea is to construct a regularizer to represent the marketing goal, and further optimize it efficiently using gradient estimation techniques.
Concretely, we use S-Learner~\cite{kunzel2019metalearners} based on Deep Neural Networks (DNNs) to predict the treatment effects of different users.
Given the estimated treatment effects, we further formulate the budget allocation problem as a multiple choice knapsack problem (MCKP)~\cite{kellerer2004multiple}, and exploit the Expected Outcome Metric method (EOM) to build an unbiased estimation (denoted as $Q$) of the marketing goal (i.e., the expected business goal under a certain budget constraint).
The $Q$ function represents the marketing effectiveness and is the exact objective we want to maximize, thus we treat $Q$ as a regularizer and integrate it into the training objective of the S-Learner model.
However, there remains one challenge in doing so: since our construction process of $Q$ involves solving a series of MCKPs using Lagrange multipliers and thus many non-differentiable operations are involved in, the automatic differentiation mechanism in modern machine learning libraries (such as TensorFlow~\cite{tensorflow2015-whitepaper} and PyTorch~\cite{paszke2019pytorch}) would not give correct gradients.
Our solution to resolve this issue is to treat $Q$ as a black-box function and exploit gradient estimation techniques such as the finite-difference method or natural evolution strategies (NES)~\cite{wierstra2014natural} to obtain the gradients.
As such, the regularizer could be effectively jointly optimized with the original learning objective of the S-Leanrer model.
This regularizer endows the S-Learner model with an ability to directly learn from the marketing goal and further maximize it, in a principled way.
Our proposed method is extensively evaluated on both offline simulations and online experiments.
Experimental results demonstrate that our proposed end-to-end training framework can achieve significantly better marketing effectiveness, in comparison with current state-of-the-arts.
We further deploy the proposed method in a large-scale short video platform to allocate cash bonuses to users to incentivize them to create more videos, and online A/B tests show our method significantly outperforms the baseline methods by at least 1.24\% more video creators under the same budget, which is a tremendous improvement over these months.
Currently, our proposed method is serving hundreds of millions of users.

\section{Related Works}\label{sec:related_work}
\subsection{Budget Allocation}
Starting from a common objective, many budget allocation methods have been proposed.
Early methods~\cite{guelman2015uplift,zhao2017uplift,zhao2019uplift} often select the amount of incentives for users heuristically after obtaining the heterogeneous treatment effects estimation.
However, lacking an explicit optimization problem formulation could make these methods less effective in maximizing the marketing goal.

A popular way of allocating budget is to follow a two-stage paradigm~\cite{zhao2019unified,makhijani2019lore,goldenberg2020free,tu2021personalized,ai2022lbcf,albert2022commerce}: in the first stage uplift models are used to predict the treatment effects, and in the second stage integer programming is invoked to find the optimal allocation.
For instance, \citet{zhao2019unified} propose to use a logit response model~\cite{phillips2021pricing} to predict treatment effects and then obtain the optimal allocation via root-finding for KKT conditions.
Later, \citet{tu2021personalized} introduce various advanced estimators (including Causal Tree~\cite{athey2016recursive}, Causal Forest~\cite{wager2018estimation} and Meta-Learners~\cite{soltys2015ensemble,kunzel2019metalearners}) to estimate the heterogeneous effects.
The second stage has also received much research attention. 
\citet{makhijani2019lore} formulate the marketing goal as a Min-Cost Flow network optimization problem for better expressivity.
More recently, \citet{ai2022lbcf,albert2022commerce} use MCKP to represent the budget allocation problem in the discrete case and develop efficient solvers based on Lagrangian duality.
Though effective, owing to the misalignment between the objectives of these two stages, solutions from these two-stage methods could be suboptimal.

Several methods have also been proposed to directly learn the optimal budget assignment policy~\cite{xiao2019model,du2019improve,goldenberg2020free,zou2020heterogeneous,zhang2021bcorle,zhou2023direct}.
\citet{xiao2019model,zhang2021bcorle} develop reinforcement learning solutions based on constrained Markov decision process to directly learn an optimal policy.
However, learning such complicated policies in a pure model-free black-box approach without exploiting the causal structures could suffer from sample inefficiency.
\citet{du2019improve,zou2020heterogeneous,goldenberg2020free} propose to directly learn the ratios between values and costs in the binary treatment setting, and treatments could be first applied to users with higher scores.
Later, a contemporary work~\cite{zhou2023direct} proves the proposed loss in~\cite{du2019improve,zou2020heterogeneous} cannot achieve the correct rank when
the loss converges.
Furthermore, the method proposed in~\cite{goldenberg2020free} is applicable only if the budget constraint is that monetary ROI$\ge0$, which is not flexible enough for many marketing campaigns in online Internet platforms.
\citet{zhou2023direct} extend the decision-focused learning idea to the multiple-treatments setting and develop a loss function to learn decision factors for MCKP solutions.
However, the proposed method relies on the assumption of diminishing marginal utility, which is often not strictly hold in practice.

\subsection{Gradient Estimation}
Estimating the gradients of a black-box function has been extensively studied for decades and widely used in the field of reinforcement learning~\cite{sutton1999policy} and adversarial attack~\cite{ilyas2018black,guo2019subspace}.
The simplest approach is the finite-difference strategy, i.e., calculating the gradients by definition.
The finite-difference strategy could give accurate gradients but might be computationally expensive if the independent variables of the black-box function have high dimensionalities.
An alternative way of gradient estimation is natural evolution strategies (NES)~\cite{wierstra2014natural}, which generate several search directions to probe local geometrics of the black-box function and estimate the gradients based on sampled search directions.
NES is generally more computationally efficient than the finite-difference strategy, but the estimated gradients are noisy.
Antithetic sampling~\cite{salimans2017evolution} could be employed to reduce the variance of NES gradient estimation.
In this paper, we present both finite-difference- and NES-based algorithms for gradient estimation.

\section{Our End-to-End Framework}\label{sec:method}

Many budget allocation methods follow a two-stage paradigm, which may lead to degraded performance due to the misalignment between the objectives of the two stages.
We reckon it can be beneficial to properly incorporate the allocation process based on the second stage into the learning objective, such that the model can directly learn to maximize the marketing goal. 
In this section, we first introduce major notations, then briefly analyze the two-stage paradigm, finally we present our end-to-end training framework.

\subsection{Notations}\label{sec:notation}
In most online Internet platforms, randomized controlled trials (RCT) could be easily deployed and thus a decent amount of RCT data could usually be collected at an acceptable cost in practice.
As such, in this paper, we focus on training models using RCT data for simplicity and leave the exploration on exploiting observational data to future works. 
We use $u_i\in\mathbb R^d$ to denote the features for $i$-th user in a certain user group.
The term ``user group'' might refer to a group of RCT training users or a group of non-RCT users whose budgets are allocated by our model.
The exact meaning of the term ``user group'' should be easily inferred from the context in this paper.
We set $K$ mutually exclusive treatments, and each treatment represents a certain amount of incentives (e.g., cash bonus, or discount levels) assigned to the user.
The treatment assigned to user $u_i$ in RCT is denoted by $t_i\in\{1,\ldots K\}$, and the user would have a binary response $y_i\in\{0,1\}$ after receiving the treatment $t_i$.
The response shall be the business goal we want to maximize. 
Meanwhile, applying treatment $t_i$ to user $u_i$ would induce a non-negative cost $z_i\ge 0$.
Each sample from an RCT dataset consists of a quadruple $(u_i, t_i, y_i, z_i)$.

\subsection{The Two-Stage Paradigm}\label{sec:method_two_stage}
As discussed, a lot of efforts have been devoted to two-stage budget allocation methods~\cite{zhao2019unified,makhijani2019lore,goldenberg2020free,tu2021personalized,ai2022lbcf,albert2022commerce}.
In this section, we briefly introduce common components in two-stage budget allocation methods.
In the first stage, two-stage methods seek to build an uplift model to predict the uplifts of responses under all possible treatments.
In principle, any uplift models could be applied in this stage, such as S-Learners~\cite{kunzel2019metalearners}, R-Learners~\cite{nie2021quasi}, Uplift Trees~\cite{rzepakowski2012decision}, Causal Forests~\cite{athey2016recursive}, GRFs~\cite{athey2019generalized}, ORFs~\cite{oprescu2019orthogonal} and LBCFs~\cite{ai2022lbcf}.
Although some tree-based uplift models like GRFs~\cite{athey2019generalized} and ORFs~\cite{oprescu2019orthogonal} are unbiased uplift estimators with theoretical convergence guarantees, the splitting criteria in their tree-growing process are not flexible enough to be easily extended and integrated with our proposed regularizers.
Thus in this paper, we restrict our attention to using S-Leaners with DNNs~\cite{krizhevsky2012imagenet} as base models, just like~\cite{zhou2023direct}.
A typical DNN S-Learner model $f$ would take the user feature $u_i$ as input and feed $u_i$ into a multi-layer perceptron (MLP) with several fully-connected layers to obtain the base hidden representation $h_i$ of the user $u_i$, and then the model $f$ would branch $2K$ headers (i.e., shadow MLPs) on the top of the base hidden representation $h_i$ to predict responses and costs for every possible treatment, where each treatment corresponds two headers (one header for response and the other header for cost).
In general, any DNN architecture could be used in this stage provided the DNN model takes $u_i$ as input and outputs two $K$-dimensional vectors for predicted responses/costs under all possible treatments, and in this paper, we use the simple architecture described above and leave the exploration on DNN architectures to future works. 
Let $\theta$ exhaustively collects learnable parameters of the above DNN S-Leaner $f$, and let $f_v\left(u_i;\theta\right)\in\mathbb R^K$ and $f_c\left(u_i;\theta\right)\in\mathbb R^K$ denote the predicted responses/costs of $f$.
The predicted response and cost of applying treatment $j$ to user $u_i$ are denoted by $f_v\left(u_i;\theta\right)_j$ and $f_c\left(u_i;\theta\right)_j$, respectively.
Given a training set with $N$ RCT users, the DNN S-Learner is trained in the first stage by minimizing the following objective:
\begin{equation}\label{eqn:loss_s_learner}
  L_{\textrm{SL}}\left(\theta\right)=-\frac{1}{N}\sum_{i=1}^N L_{\textrm{CE}}\left(f_v\left(u_i;\theta\right)_{t_i}, y_i\right)+L_{\textrm{C}}\left(f_c\left(u_i;\theta\right)_{t_i}, z_i\right),
\end{equation}
where $L_{\textrm{CE}}(\cdot,\cdot)$ is the cross entropy loss for the binary response, and $L_{\textrm{C}}(\cdot,\cdot)$ is a suitable loss for the cost target.
The formulation of $L_{\textrm{C}}(\cdot,\cdot)$ could vary from one dataset to another.
For example, on a dataset where the cost label $z_i$ is categorical, it could be reasonable to use a cross entropy-loss for $L_{\textrm{C}}(\cdot,\cdot)$, while on another dataset where the cost label $z_i$ is continuous, a regression loss would be more appropriate for $L_{\textrm{C}}(\cdot,\cdot)$.
After minimizing the objective $L_{\textrm{SL}}\left(\theta\right)$ in Equation~\eqref{eqn:loss_s_learner}, given a feature vector of a user, the well-trained DNN S-Leaner model should predict unbiased estimations of expected responses/costs of this user under all possible treatments.

In the second stage, two-stage methods usually keep the uplift model $f$ fixed and use a multiple choice knapsack problem (MCKP)~\cite{kellerer2004multiple} formulation to maximize the marketing goal under the budget constraint.
%To construct an MCKP, the total budget for all users, the costs, and the expected responses of all users under all possible treatments are required.
Suppose the total budget for a group of $M$ users is set to $T$. 
The expected response of $u_i$ under treatment $j$ is denoted by $v_{ij}\in[0, 1]$, and $v_{ij}$ could be predicted by the well-trained DNN S-Learner from the first stage: $v_{ij}=f_v\left(u_i;\theta\right)_j$.
Let $c_{ij}\ge 0$ denotes the expected cost of $u_i$ under treatment $j$, and $c_{ij}$ could also be similarily predicted by $c_{ij}=f_c\left(u_i;\theta\right)_j$.
Using the above coefficients and constants, two-stage methods typically construct the following MCKP to find the optimal personalized treatment assignment solution $x\in\{0,1\}^{M\times K}$ which could maximize the expected marketing goal under a given budget constraint:
\begin{equation}\label{eqn:mckp}
\begin{aligned}
  \max_{x} \quad & \sum_{i=1}^M\sum_{j=1}^K v_{ij}x_{ij} \\
  \textrm{s.t.} \quad & \sum_{i=1}^M\sum_{j=1}^K c_{ij}x_{ij}\le T\\
    & \sum_{j=1}^Kx_{ij}=1,\quad i=1,2,\ldots,M   \\
    & x_{ij}\in\{0,1\},\quad i=1,2,\ldots,M,\quad j=1,2,\ldots,K,
\end{aligned}
\end{equation}
where the $M\times K$ matrix $x$ represents the budget allocation solution for $M$ users.
All entries in $x$ are either 0 or 1.
Each row of $x$ has exactly one entry with value 1, and $x_{ij}=1$ means user $u_i$ should be assigned with treatment $j$.
Although MCKP is proven to be NP-Hard~\cite{kellerer2004multiple}, in practice we can approximately solve it by first making linear relaxation to get rid of the integer constraints and then converting it to the dual problem~\cite{ai2022lbcf,zhou2023direct}.
Specifically, the dual problem of MCKP has only one scalar variable $\alpha$ under a constraint that $\alpha\ge 0$~\cite{kellerer2004multiple}, and the objective of dual problem is given by
\begin{equation}\label{eqn:mckp_dual}
  \begin{aligned}
    \min_{\alpha\ge 0} \quad & \alpha T+\sum_{i=1}^M\max_j\left(v_{ij}-\alpha c_{ij}\right).
  \end{aligned}
\end{equation}
The dual problem~\eqref{eqn:mckp_dual} could be efficiently solved via bisection methods such as Dyer-Zemel algorithm~\cite{zemel1984n} or dual gradient descent~\cite{ai2022lbcf,zhou2023direct}.
Once the optimal dual solution $\alpha^*$ is obtained, the optimal solution $x^*$ of the primal MCKP~\eqref{eqn:mckp} could be given by KKT conditions:
\begin{equation}\label{eqn:mckp_primal_dual_optimal}
  x_{ij}^*=\begin{cases}
    1, & \text{if } j = \argmax_k v_{ik}-\alpha^*\cdot c_{ik} \\
    0, & \text{otherwise.}
  \end{cases}
\end{equation}
Interested readers can read the insightful paper~\cite{kellerer2004multiple} for more details about solving MCKPs.

\subsection{Marketing Goal as a Regularizer}\label{sec:method_regularizer}
As discussed in Section~\ref{sec:introduction}, our core idea for end-to-end training is to construct a regularizer to represent the marketing goal (i.e., the expected response under a certain budget constraint) and optimize it efficiently using gradient estimation techniques.
Suppose we have a batch of $B$ RCT users from the training set (we set the batch size to $B=10,000$ in all experiments), the S-Learner model could predict responses for each user and we shall stack their predicted responses into a $B\times K$ matrix $v$ to simplify the notations, and the costs of applying arbitrary treatments to these users could also be stacked into a $B\times K$ matrix $c$ in a similar way.
In this section, we present our solution to design a function $Q(v,c)$ to map the predicted response matrix $v$ and cost matrix $c$ to the marketing goal and integrate $Q(v,c)$ into the training objective of the S-Leaner model as a regularizer.

\begin{figure}[t]
  \centering
  \includegraphics[width=.8\linewidth]{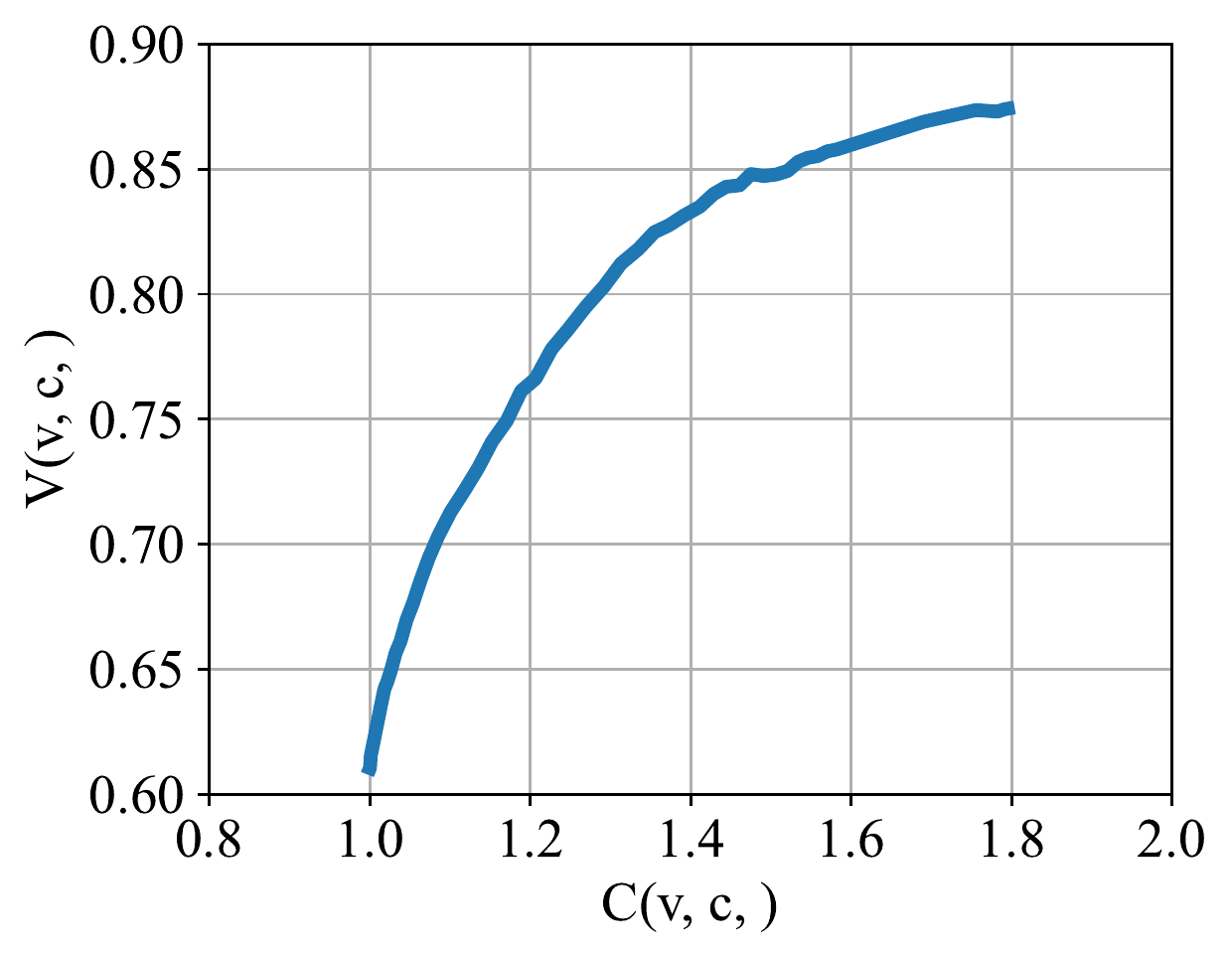}
  \caption{Example of sensitiveness curve between the empirical business goal (y-axis) and empirical cost (x-axis) in the marketing data.}
  \Description{EExample of sensitiveness curve between the empirical business goal (y-axis) and empirical cost (x-axis) in the marketing data}
  \label{fig:cost_value}
\end{figure}

We use Expected Outcome Metric (EOM) method~\cite{zhao2017uplift,ai2022lbcf,zhou2023direct} to construct $Q(v,c)$.
Concretely, starting from any dual feasible solution $\alpha\ge 0$, we can obtain the corresponding budget allocation solution $x$ via KKT conditions in Equation~\eqref{eqn:mckp_primal_dual_optimal}. 
Since $\alpha$ is the Lagrangian multiplier associated with the budget constraint in the primal MCKP~\eqref{eqn:mckp}, we know from complementary slackness conditions that $x$ must be the primal optimal solution for another MCKP in which the total budget $T$ is set to be $T=\sum_{i=1}^M\sum_{j=1}^K c_{ij}x_{ij}$ while other coefficients $c$ and $v$ remain unchanged.
In other words, given any $\alpha\ge 0$, the corresponding $x$ is the optimal budget allocation solution under a certain total budget constraint.
Using the budget allocation solution $x$, we can find the set of user indices (denoted by $\mathcal S_\alpha$) in the $B$ RCT users whose treatment on the RCT data is equal to the assigned treatment in $x$:
\begin{equation}\label{eqn:same_treatment}
  \mathcal S_\alpha = \left\{i\,\middle |\,t_i=\argmax_j x_{ij},\,\, i=1,2,\ldots,B\right\},
\end{equation}
where $t_i$ is the treatment applied to user $u_i$ in the RCT data.
Based on the set $\mathcal S_\alpha$, the per-capita response $V\left(v, c,\alpha\right)\in\mathbb R_+$ and per-capita cost $C\left(v, c,\alpha\right)\in\mathbb R_+$ could be empirically estimated by:
\begin{equation}\label{eqn:V_and_D}
  \begin{aligned}
  V\left(v, c, \alpha\right) &= \frac{1}{\left\lvert\mathcal S_\alpha\right\rvert}\sum_{i\in\mathcal S_\alpha} y_i \\
  C\left(v, c, \alpha\right) &= \frac{1}{\left\lvert\mathcal S_\alpha\right\rvert}\sum_{i\in\mathcal S_\alpha} z_i, 
  \end{aligned}
\end{equation}
where $y_i\in\{0,1\}$ and $z_i\in\mathbb R_+$ denote the response and cost of user $u_i$ in the RCT data respectively, and $\left\lvert\mathcal S_\alpha\right\rvert$ represents the cardinality of set $\mathcal S_\alpha$.

\begin{algorithm}[th]
  \caption{Evaluate $Q(v,c)$ given $v$ and $c$}\label{alg:Q}
  \begin{algorithmic}[1]
    \State {\bfseries Input:} the predicted response/cost matrix $v$/$c$ of a group of $B$ users; the total budget $T$; the maximum number of iterations $I$ in bisection; the maximum $\alpha_{\textrm{max}}$ in bisection; the tolerance $\epsilon$ of per-capita cost in bisection.
    \State {\bfseries Output:} the expected per-capita response $Q(v,c)$ under the optimal budget allocation solution.
    \State Initialize $t\gets 1$, $\alpha_l\gets 0$, $\alpha_h\gets\alpha_{\textrm{max}}$.
    \While{$t\le I$}

    \State // Evaluate the middle point in the search interval.
    \State $\alpha_m\gets\frac{\alpha_l+\alpha_h}{2}$
    \State Recover the primal solution $x_m$ corresponing to $\alpha_m$ using KKT conditions in Equation~\eqref{eqn:mckp_primal_dual_optimal}.
    \State Based on $x_m$, evaluate empirical per-capita response $V\left(v,c,\alpha_m\right)$ and empirical per-capita cost $C\left(v,c, \alpha_m\right)$ using Equation~\eqref{eqn:same_treatment} and Equation~\eqref{eqn:V_and_D}.
    \State

    \State // Check if we could break early.
    \State Calculate the per-capita cost error: $e\gets\left\lvert C\left(v,c,\alpha_m\right)-\frac{T}{B}\right\rvert$.
    \If{$e\le\epsilon$}
      \State break
    \EndIf
    \State

    \State // Update search interval.
    \If{$C\left(v,c,\alpha_m\right)>\frac{T}{B}$}
      % empirical cost too high, turn down the budget, increase alpha
      \State // Empirical cost is too high, we should turn down the budget in MCKP by increasing $\alpha$.
      \State $\alpha_l\gets\alpha_m$
    \Else
      \State // Empirical cost is too low, we should turn up the budget in MCKP by decreasing $\alpha$.
      \State $\alpha_h\gets\alpha_m$
    \EndIf
    \State 

    \State // Update step counter.
    \State $t\gets t+1$
    \EndWhile
 
    \State {\bfseries return} the final empirical per-capita response $V\left(v, c,\alpha_m\right)$.
  \end{algorithmic}
\end{algorithm}

As discussed, the obtained $x$ is the optimal budget allocation solution for these $B$ users under a certain budget constraint, and thus we could tune the total budget by adjusting the value of dual feasible solution $\alpha$~\cite{kellerer2004multiple}.
As such, we could start from $\alpha=0$ and gradually increase its value, and the total budget in MCKP would be accordingly decreased.
In this process, we could plot empirical per-capita cost $C\left(v, c,\cdot\right)$ and empirical per-capita response $V\left(v, c,\cdot\right)$ in the same figure~\footnote{Note this curve might not be strictly monotonically increasing since the empirical estimation process is noisy.} to illustrate the users' expected responses under different costs, and an example figure is shown in Figure~\ref{fig:cost_value}.
Based on this curve, intuitively we can use bisection search to find a suitable $\alpha'$ such that the empirical per-capita cost $C\left(v, c,\alpha'\right)\in\mathbb R_+$ equals the given per-capita budget $\frac{T}{\left\lvert\mathcal S_\alpha\right\rvert}$ in the original MCKP, and define $Q(v,c)$ as the empirical per-capita response at that $\alpha'$: $Q(v,c)=V\left(v,c,\alpha'\right)$.
All details of evaluating $Q(v,c)$ given $v$ and $c$ are summarized in Algorithm~\ref{alg:Q}.

The bisection process in Algorithm~\ref{alg:Q} could also be extended to $n$-ary search by adding more middle points in the search interval in each iteration.
Although the bisection method and the $n$-ary search method share the same asymptotic time complexity, in modern GPU devices the $n$-ary search method could usually run faster if the implementation is properly vectorized. 

Given a batch of training data with $B$ RCT users, instead of optimizing the original objective $L_{\textrm{SL}}\left(\theta\right)$ in Equation~\eqref{eqn:loss_s_learner} for S-Learner, we could directly optimize the marketing effectiveness by minimizing the following regularized objective $L_R(\theta)$ in this batch:
\begin{equation}\label{eqn:loss_regularized}
  L_R(\theta)=L_{\textrm{SL}}(\theta)-\lambda\cdot Q(v,c),
\end{equation}
where $L_{\textrm{SL}}(\theta)$ is the original loss in Equation~\eqref{eqn:loss_s_learner} on this batch, and $\lambda>0$ is a scaling factor that balances the importance of the original objective $L_{\textrm{SL}}\left(\theta\right)$ and the regularizer $Q(v,c)$.

If the gradient $\frac{\partial Q(v,c)}{\partial v}$ and $\frac{\partial Q(v,c)}{\partial c}$ could be obtained (see Section~\ref{sec:method_grad_estimation}), we can integrate the gradient into the back-propagation process (see Section~\ref{sec:method_training}).
As such, the regularizer $Q(v,c)$ could be jointly optimized with the original learning objective $L_{\textrm{SL}}(\theta)$ and the whole objective $L_R(\theta)$ in Equation~\eqref{eqn:loss_regularized} could be efficiently solved in an end-to-end flavor.
This endows the trained model with an ability to directly learn from the marketing goal and further maximize it, in a principled way.

\subsection{Gradient Estimation}\label{sec:method_grad_estimation}
As shown in Algorithm~\ref{alg:Q}, the construction of regularizer $Q(v,c)$ involves many non-differentiable operations.
Thus if we na\"ively implement Algorithm~\ref{alg:Q} as a computational graph in modern machine learning libraries such as TensorFlow~\cite{tensorflow2015-whitepaper} and PyTorch~\cite{paszke2019pytorch}, the automatic differentiation mechanism would not generate correct gradients.
In this section, we present our solution to overcome this challenge and calculate the (estimation of) gradient $\frac{\partial Q(v,c)}{\partial v}$ and $\frac{\partial Q(v,c)}{\partial c}$.
We treat $Q(v,c)$ as a black-box function, and use two strategies to estimate the gradients: 1) finite-difference strategy in Section~\ref{sec:method_grad_estimation_fd}, and 2) natural evolution strategies (NES)~\cite{wierstra2014natural} in Section~\ref{sec:method_grad_estimation_nes}.
For clarity of presentation, in this section, we would present the detailed solution for estimating $\frac{\partial Q(v,c)}{\partial v}$ only, and estimating for $\frac{\partial Q(v,c)}{\partial c}$ could be similarly performed.

\subsubsection{Finite-Difference Strategy}\label{sec:method_grad_estimation_fd}
Finite-difference strategy is commonly used in the field of black-box adversarial attack~\cite{chen2017zoo,tu2019autozoom} to estimate the gradient of a complicated black-box function.
Specifically, given predicted responses $v\in [0,1]^{B\times K}$ of a batch of $B$ RCT training users, the $l$-th element in the gradient matrix $\frac{\partial Q(v,c)}{\partial v}\in\mathbb{R}^{B\times K}$ could be estimated by:
\begin{equation}\label{eqn:fd}
  \left(\frac{\partial Q(v,c)}{\partial v}\right)_l\approx\frac{Q\left(v+h\cdot e_l,c\right)-Q\left(v-h\cdot e_l,c\right)}{2h},
\end{equation}
where $l=1,2,\ldots,B\times K$, $h$ is a small positive constant to perturb the input of the $Q$ function, and $e_l\in \{0,1\}^{B\times K}$ is the standard basis matrix with only the $l$-th component as 1.
We use the symmetric difference quotient~\cite{peter2014calculus} to improve the numerical stability.

Using Equation~\eqref{eqn:fd}, theoretically, we could estimate each entry of the matrix $\frac{\partial Q(v,c)}{\partial v}$ one by one, and finally stack results of all entries together to obtain the estimation of the entire gradient matrix $\frac{\partial Q(v,c)}{\partial v}$, at the cost of evaluating the $Q$ function $2\times B\times K$ times.
However, the forward process of function $Q$ is sophisticated, and thus evaluating the $Q$ function for $2\times B\times K$ times could be computationally intensive in practice.
To accelerate the finite-difference process, we use a strategy somewhat analogous to a coordinate descent: randomly masking some entries.
Concretely, in each training iteration, instead of evaluating all $B\times K$ entries of $\frac{\partial Q(v,c)}{\partial v}$, we randomly select $F'\ll B\times K$ entries and evaluate these entries of $\frac{\partial Q(v,c)}{\partial v}$ using Equation~\eqref{eqn:fd} while other entries are set to zero.
In practice, we could adjust the hyper-parameter $F'$ to achieve a sensible trade-off between computational efficiency and the accuracy of the gradient estimation.

\subsubsection{Natural Evolution Strategy}\label{sec:method_grad_estimation_nes}
NES is a popular derivative-free numerical optimization algorithm and it is commonly used in the field of black-box adversarial attack~\cite{ilyas2018black,ilyas2018prior,guo2019subspace}.
Suppose a random variable $v'\in\mathbb{R}^{B\times K}$ follows an isotropic normal distribution with $v$ as the center and $\sigma^2I$ as the covariance matrix: $v'\sim\mathcal{N}\left(v,\sigma^2I\right)$, where $\sigma$ is a small positive constant and $I$ is a $\left(B\times K\right)$-by-$\left(B\times K\right)$ identity matrix.
NES seeks to maximize the expected value of function $Q$ under the distribution of $v'$ by estimating the gradient using the ``log derivative trick''\footnote{Similar trick is also used in policy gradient algorithms in reinforcement learning.}:
\begin{equation}\label{eqn:nes}
  \begin{aligned}
  \frac{\partial}{\partial v}\mathbb{E}_{v'}\left[Q\left(v',c\right)\right]&=\frac{\partial}{\partial v}\int Q\left(v',c\right)\pi\left(v'\middle|\,v\right)\,\textrm{d}v'\\ 
    &=\int Q\left(v',c\right)\frac{\partial}{\partial v}\pi\left(v'\middle|\,v\right)\,\textrm{d}v'\\
    &=\int Q\left(v',c\right)\frac{\pi\left(v'\middle|\,v\right)}{\pi\left(v'\middle|\,v\right)}\frac{\partial}{\partial v}\pi\left(v'\middle|\,v\right)\,\textrm{d}v'\\
    &=\int Q\left(v',c\right)\pi\left(v'\middle|\,v\right)\frac{\partial}{\partial v}\log\left(\pi\left(v'\middle|\,v\right)\right)\,\textrm{d}v'\\
    &=\mathbb{E}_{v'}\left[Q\left(v',c\right)\frac{\partial}{\partial v}\log\left(\pi\left(v'\middle|\,v\right)\right)\right],
  \end{aligned}
\end{equation}
where $\pi\left(v'\middle|\,v\right)$ is the probability density function of the random variable $v'$.
If the variance $\sigma^2$ is small enough, the expected value of $Q$ in Equation~\eqref{eqn:nes} should be close to $Q(v,c)$, and thus we could use NES to estimate the gradient $\frac{\partial Q(v,c)}{\partial v}$:
\begin{equation}\label{eqn:nes2}
  \begin{aligned}
  \frac{\partial Q(v,c)}{\partial v}&\approx\frac{\partial}{\partial v}\mathbb{E}_{v'}\left[Q\left(v',c\right)\right]\\
   &=\mathbb{E}_{v'}\left[Q\left(v',c\right)\frac{\partial}{\partial v}\log\left(\pi\left(v'\middle|\,v\right)\right)\right]\\
   &\approx \frac{1}{\sigma N'}\sum_{i=1}^{N'}\delta_i\cdot Q\left(v+\sigma\delta_i,c\right),
  \end{aligned}
\end{equation}
where in the last equation we use empirical average value to estimate the expectation, $\delta_i\in\mathbb{R}^{B\times K}$ is a sample from the standard normal distribution in the $B\times K$ dimensional vector space, and $N'$ is the number of sampling directions.
Using Equation~\eqref{eqn:nes2}, we could obtain an unbiased estimation of the gradient $\frac{\partial Q(v,c)}{\partial v}$ in a derivative-free fashion.
The hyper-parameter $N'$ controls the trade-off between the computational efficiency and accuracy of NES.

\begin{algorithm}[th]
  \caption{Train DNN model end-to-end for budget allocation}\label{alg:train}
  \begin{algorithmic}[1]
    \State {\bfseries Input:} a training dataset consists of RCT users; initialized DNN model parameter $\theta$; 
    \State {\bfseries Output:} a well-trained DNN model for budget allocation
    \While{not converged}
    \State Get one batch of training data from the training set.
    \State Run forward-propagation and get predicted results $v$ and $c$.
    \State Estimate $\frac{\partial Q(v,c)}{\partial v}$ and $\frac{\partial Q(v,c)}{\partial c}$ using finite-difference strategy (Equation~\eqref{eqn:fd}) or NES (Equation~\eqref{eqn:nes2}).
    \State Calculate the surrogate loss $L_S(\theta)$ using Equation~\eqref{eqn:loss_surrogate}.
    \State Run back-propagation to obtain $\frac{\partial L_S(\theta)}{\partial\theta}$.
    \State Run one gradient descent step to update $\theta$ based on $\frac{\partial L_S(\theta)}{\partial\theta}$.
    \EndWhile
 
    \State {\bfseries return} the well-trained DNN model $\theta$.
  \end{algorithmic}
\end{algorithm}

Like other Monte Carlo methods, NES also suffers from high variance when estimating a vector in a high dimensional space.
To reduce the variance, we employ antithetic sampling~\cite{salimans2017evolution} to generate $N'$ sampling directions: we first sample $\frac{N'}{2}$ i.i.d. Gaussian noises as $\delta_i,\, i=1,2,\ldots,\frac{N'}{2}$, and then set the remaining $\frac{N'}{2}$ directions by flipping the signs of previous $\frac{N'}{2}$ directions: $\delta_i=-\delta_{N'-i+1},\, i=\frac{N'}{2}+1,\frac{N'}{2}+2,\ldots,N'$.
This sampling strategy has been empirically verified to be effective in many previous works~\cite{ilyas2018black,ilyas2018prior,guo2019subspace}.

\subsection{Training}\label{sec:method_training}
%As discussed in Section~\ref{sec:method_regularizer}, we could optimize the regularized objective $L_R(\theta)$ in Equation~\eqref{eqn:loss_regularized} to endow the DNN model the ability of directly learning from the marketing goal.
In this section, we present our solution to train the DNN model to maximize the marketing goal, by optimizing the regularized objective $L_R(\theta)$ in Equation~\eqref{eqn:loss_regularized}.
Common gradient descent based optimizers such as SGD or Adam~\cite{kingma2015adam} would require the gradients w.r.t. DNN model parameter (i.e., $\frac{\partial Q(v,c)}{\partial\theta}$) to update $\theta$, while in Section~\ref{sec:method_grad_estimation} we only have $\frac{\partial Q(v,c)}{\partial v}$ and $\frac{\partial Q(v,c)}{\partial c}$.
Thus, a back-propagation step which converts $\frac{\partial Q(v,c)}{\partial v}$ and $\frac{\partial Q(v,c)}{\partial c}$ to $\frac{\partial Q(v,c)}{\partial\theta}$ is required.
To this end, we design the following surrogate loss $L_S(\theta)$:
\begin{equation}\label{eqn:loss_surrogate}
  \begin{split}
    L_S(\theta)=L_{\textrm{CE}}(\theta)-\lambda\cdot\Tr\left[v^\mathrm{T}\texttt{stop\_gradient}\left(\frac{\partial Q(v,c)}{\partial v}\right)\right]\\
    -\lambda\cdot\Tr\left[c^\mathrm{T}\texttt{stop\_gradient}\left(\frac{\partial Q(v,c)}{\partial c}\right)\right],
  \end{split}
\end{equation} 
where $\lambda>0$ is a scaling factor that balances the importance of the original objective and the regularizer, $\Tr[\cdot]$ is the matrix trace operator, $v^\mathrm{T}$ and $c^\mathrm{T}$ is the transpose of matrix $v$ and $c$ respectively, and $\texttt{stop\_gradient}(\cdot)$ is the operator to treat a certain variable as a constant and avoid back-propagation through it.\footnote{In TensorFlow this could be achieved via $\texttt{tf.stop\_gradient()}$, and in PyTorch this could be achieved by \texttt{tensor.detach()}.}
Unlike the regularizer part of $L_R(\theta)$ which is cumbersome to handle, the regularizer part of $L_S(\theta)$ is a simple linear combination of elements in matrix $v$ and $c$, which are output values from the last layer of the DNN model.
As such, the gradient of $L_S(\theta)$ w.r.t. $\theta$ could be easily obtained by running a standard back-propagation over the DNN model.
Furthermore, it could be verified that $L_S(\theta)$ and $L_R(\theta)$ shares the same first order gradient: $\frac{\partial L_S(\theta)}{\partial\theta}=\frac{\partial L_R(\theta)}{\partial\theta}$.
Thus, with the help of the surrogate loss $L_S(\theta)$, we could effectively optimize the original regularized objective $L_R(\theta)$, and the well-trained DNN model shall learn to maximize the marketing goal in an end-to-end manner.
Algorithm~\ref{alg:train} summarizes all details for training.

\section{Experimental Results}\label{sec:experiment}
%We extensively evaluate our proposed method on various datasets and metrics.
%We also design several experiments on a synthetic dataset to illustrate the basic motivation of our method.

\subsection{Datasets}\label{sec:experiment_datasets}
We evaluate the efficacy of our method in three different datasets: 
\begin{itemize}
  \item Synthetic dataset. 
    To elaborate on key components of our idea (i.e., EOM and gradient estimation), we generate a synthetic dataset consisting of 10,000 RCT users.
    We set four different treatments (i.e., $K=4$), and the treatment $t_i$ applied to the $i$-th user in RCT is randomly sampled from these four treatments with equal probabilities.
    We assume the $i$-th user has a four-dimensional ground truth vector $v^{\textrm{gt}}_i\in [0,1]^4$, where the $j$-th element in $v^{\textrm{gt}}_i$ (denoted by $v^{\textrm{gt}}_{ij}$) represents the probability of $y_i=1$ when treatment $j$ is applied to that user.
    We use the following three steps to generate $v^{\textrm{gt}}_i$: 1) the first element of $v^{\textrm{gt}}_i$ (i.e., $v^{\textrm{gt}}_{i1}$) is sampled from the uniform distribution in $[0, 0.1]$, 2) to determine the remaining elements of $v^{\textrm{gt}}_i$, we sample three i.i.d. numbers from the uniform distribution in $[0, 0.2]$ and sort these three numbers in descending order (denoted by $0.2\ge k_1\ge k_2\ge k_3\ge 0$), and 2) these three numbers are used as incremental values of $v^{\textrm{gt}}$: let $v^{\textrm{gt}}_{ij}=v^{\textrm{gt}}_{i(j-1)}+k_{j-1}$, where $j=2,3,4$.
    Once the four-dimensional ground truth vector $v^{\textrm{gt}}_{i}$ is generated, we get the $t_i$-th element of the $v^{\textrm{gt}}_{i}$ (denoted by $v^{\textrm{gt}}_{i(t_i)}$) and sample the binary response $y_i$ of the $i$-th user from the Bernoulli distribution with probability $v^{\textrm{gt}}_{i(t_i)}$.
    We assume the cost of applying the $j$-th treatment to any user is $j$.
    In other words, we set the $\left(i,j\right)$-th element in ground-truth cost matrix $c^{\textrm{gt}}$ to be $c^{\textrm{gt}}_{ij}=j$, where $j=1,2,3,4$. 
    The synthetic dataset is designed for illustrating the effectiveness of EOM and gradient estimation, thus the features of users are omitted in this dataset.
  \item CRITEO-UPLIFT v2~\cite{diemert2018large}. 
    This publicly available dataset is designed for evaluating models for uplift modeling.
    This dataset contains 13.9 million samples which are collected from an RCT.
    Each sample has 12 dense features, one binary treatment indicator, and two binary labels: visit and conversion.
    In this dataset, treatment is defined as whether the user is targeted by advertising or not, and labels are defined as positive if the user visited/converted on the advertiser's website during a test period of two weeks.
    To evaluate different budget allocation methods, we follow~\cite{zhou2023direct} and use the visit/conversion label as the cost/value respectively.
    We compare the performance of our proposed method with state-of-the-arts in CRITEO-UPLIFT v2 to demonstrate the effectiveness of our method.
  \item KUAISHOU-PRODUCE-COIN. 
    Produce Coin is a common marketing campaign to incentivize short video creators to upload more short videos in Kuaishou.
    In this campaign, we provide a prized task to each short video creator on the Kuaishou platform.
    Concretely, in each task, if a creator uploads a video within 24 hours, he/she will receive a certain amount of coins as a reward.
    Each creator could see the number of coins he/she could possibly obtain, and if the number seems attractive enough the creator might finish the task to collect these coins.
    We manually select 30 distinct levels of coins (e.g., 1 coin, 2 coins, $\ldots$), and each creator will be assigned one of these 30 levels.
    The amount of coins could be personalized to maximize the total amount of video uploads under the constraint that the total amount of coins does not exceed a certain limit.
    In this dataset, we define treatment as the level of coins, thus we have $K=30$ treatments.
    For the user (i.e., creator) $u_i$, the response $y_i$ would be $y_i=1$ if the user has successfully completed the task, and otherwise, we set $y_i=0$.
    The cost of applying a treatment to $u_i$ would be defined as the real number of coins sent to $u_i$: if $y_i=1$ we set the cost to the number of coins corresponding to the assigned treatment, and if $y_i=0$ we set the cost to zero.
    To collect this dataset we conduct an RCT for one week.
    In this RCT, we randomly sample one reward level from predefined 30 levels with equal probabilities for each user as the treatment and collect the response/cost label from the logs.
    This dataset contains 82.4 million samples, and we deploy several models trained on this dataset in the Kuaishou platform to evaluate the online performance of different methods.
    To protect data privacy, we normalize the per-capita response/cost and report the normalized values on all tables and figures.
\end{itemize}

\subsection{Evaluation Metrics}\label{sec:experiment_metrics}
We use the following evaluation metrics to evaluate and compare the performance of different methods:
\begin{itemize}
  \item AUCC (Area Under the Cost Curve) in~\cite{du2019improve}.
    AUCC is commonly used in existing literatures~\cite{du2019improve,ai2022lbcf,zhou2023direct} to evaluate the ranking performance of uplift models in the two-treatment setting.
    Interested readers can check the original paper~\cite{du2019improve} for more details about the evaluation process of AUCC.
    In this paper, we use AUCC to compare the performance of different methods in CRITEO-UPLIFT v2.
  \item EOM (Expected Outcome Metric).
    EOM or similar metric is also commonly used in existing literatures~\cite{ai2022lbcf,zhou2023direct} to empirically estimate the expected outcome (e.g., per-capita response or per-capita cost) for arbitrary budget allocation policy.
    EOM can make an unbiased estimation of any outcome, provided that outcome is calculable given a budget allocation solution on RCT data.
    Thus, EOM is more flexible than AUCC in practice.
    In this paper, we use EOM in the synthetic dataset and the KUAISHOU-PRODUCE-COIN dataset to evaluate different budget allocation methods, and technical details of EOM have been introduced in Equation~\eqref{eqn:same_treatment} and Equation~\eqref{eqn:V_and_D} in Section~\ref{sec:method_regularizer}.
\end{itemize}

\subsection{Implementation Details}\label{sec:experiment_details}
Implementation details on different datasets are shown as follows:
\begin{itemize}
  \item The synthetic dataset.
    To illustrate the basic idea of EOM and gradient estimation in our proposed method, we first generate a cost matrix $c\in\mathbb{R}_+^{10,000\times5}$ and a value matrix $v\in[0,1]^{10,000\times5}$ as the starting point, and then use the Adam optimizer with a learning rate 0.005 to optimize $v$ by maximizing $Q(v, c)$ for 100 gradient ascent steps.
    For the starting point of cost matrix $c$, we directly use the ground-truth cost matrix $c=c^{\textrm{gt}}$ and keep it fixed during gradient ascent.
    For the starting point of value matrix $v$, we use the same procedure as described in Section~\ref{sec:experiment_datasets} to re-generate a random matrix $v$ which satisfies the following conditions for all $i$: 1) $0\le v_{i1}\le v_{i2} \le v_{i3} \le v_{i4}$, and 2) $v_{i2}-v_{i1}\ge v_{i3}-v_{i2}\ge v_{i4}-v_{i3} \ge 0$.
    We set the per-capita budget to 2.0 for evaluating the $Q$ function.
    Since features of users are omitted in the synthetic dataset, we could not expect any generalization. 
    Thus, we treat all 10,000 samples as training samples and report per-capita response and per-capita cost estimated by EOM on all training samples.
  \item CRITEO-UPLIFT v2.
    We compare our proposed method with the following baseline methods in terms of AUCC:
    \begin{itemize}
      \item TSM-SL. 
        The baseline two-stage method in many existing literatures~\cite{ai2022lbcf,zhao2019unified,zhou2023direct}.
        In the first stage, a DNN-based S-Learner model is exploited to predict the conversion uplifts and visit uplifts for individuals.
        In the second stage, an MCKP formulation could be used to find the optimal budget allocation solution.
        However for AUCC, the explicit budget allocation solution is not required, and we simply rank different individuals by the ratio between the conversion uplifts and visit uplifts and use the rank to compute AUCC.
      \item Direct Rank~\cite{du2019improve}.
        \citet{du2019improve} propose a direct rank method to directly learn the ratios between values and costs and use the ratios to rank individuals.
      \item DRP~\cite{zhou2023direct}.
        \citet{zhou2023direct} design several decision factors for MCKP solutions, and propose a surrogate loss function to directly learn these decision factors.
    \end{itemize}
    Following~\cite{zhou2023direct}, we randomly sample 70\% samples from the dataset as the training set, and the remaining samples are used as the test set.
    We report the performance of different methods in the test set, and for baseline methods, we directly cite results from~\cite{zhou2023direct}.
    For our proposed method, we use a DNN architecture described in Section~\ref{sec:method_two_stage}: the base DNN is a 4-layer MLP (hidden sizes 512-256-128-64), and each header DNN is a 2-layer MLP (hidden sizes 32-1).
    We insert a ReLU activation layer after each fully connected layer in both base DNN and header DNN, except the final output layer in header DNN.
    We also insert a Batch Normalization layer~\cite{ioffe2015batch} after the activation layer for faster convergence.
    The value of $\lambda$ in the surrogate loss $L_S(\theta)$ in Equation~\eqref{eqn:loss_surrogate} is set to 200.
    For the finite-difference strategy, we set $F'=4,000$ and $h=0.0003$.
    For NES, we set $N'=2,000$ and $\sigma=0.001$.
    The MSRA initialization~\cite{he2015delving} is invoked to initialize all weight matrices in our model.
    The randomly initialized DNN S-Learner model is trained using the Adam optimizer with a learning rate of 0.0001 for 10 epochs.
    In each training step, the per-capita budget for evaluating the $Q$ function is randomly sampled from a uniform distribution in $[0.039, 0.047]$.
    Following~\cite{zhou2023direct}, we run our proposed method 20 times and report the mean and variance of results.
  \item KUAISHOU-PRODUCE-COIN.
    To demonstrate the effectiveness of our proposed training framework in real-world scenarios, we conduct both offline evaluation and online evaluation to compare our proposed method and TSM-SL.
    The offline experiment and online experiment share the following settings:
    \begin{itemize}
      \item The architecture for base DNN header DNN is kept the same as in the CRITEO-UPLIFT v2 dataset.
      \item We train the DNN S-Learner model using the Adam optimizer with a learning rate of 0.0003 for 20 epochs.
          To make a fair comparison, we use identical feature engineering/DNN architecture/training policies for our proposed method and TSM-SL.
      \item In each training step, the per-capita budget for evaluating the $Q$ function is randomly sampled from a uniform distribution in $[1.0, 1.5]$.
    \end{itemize}
    Furthermore, in the offline experiment:
    \begin{itemize}
      \item We randomly sample 80\% of users from the KUAISHOU-PRODUCE-COIN dataset as the training set and the remaining 20\% is used as the test set.
      \item We employ EOM and bisection to find the value of per-capita response when per-capita cost equals 1.2, and this metric is used to measure the performance of different methods.
    \end{itemize}
    In the online experiment:
    \begin{itemize}
      \item All samples in the KUAISHOU-PRODUCE-COIN dataset are used for training and the obtained well-trained models are deployed in an online A/B test in the Kuaishou platform to compare the efficacy of their budget allocation policies in the real world.
      \item The online A/B test is conducted for one week. 
        There are 14.2 million users in the online A/B test, and these users are disjoint for users in the KUAISHOU-PRODUCE-COIN dataset.
        We randomly partition 14.2 million users into two equally sized groups, one group for our proposed method and the other group for TSM-SL.
        For each group, the reported per-capita response/cost is directly computed over the entire 7.1 million users, and EOM is not employed.
    \end{itemize}
\end{itemize}

\begin{figure}[thbp]
  \centering
  \begin{subfigure}[b]{0.48\linewidth}
   \includegraphics[width=1.0\linewidth]{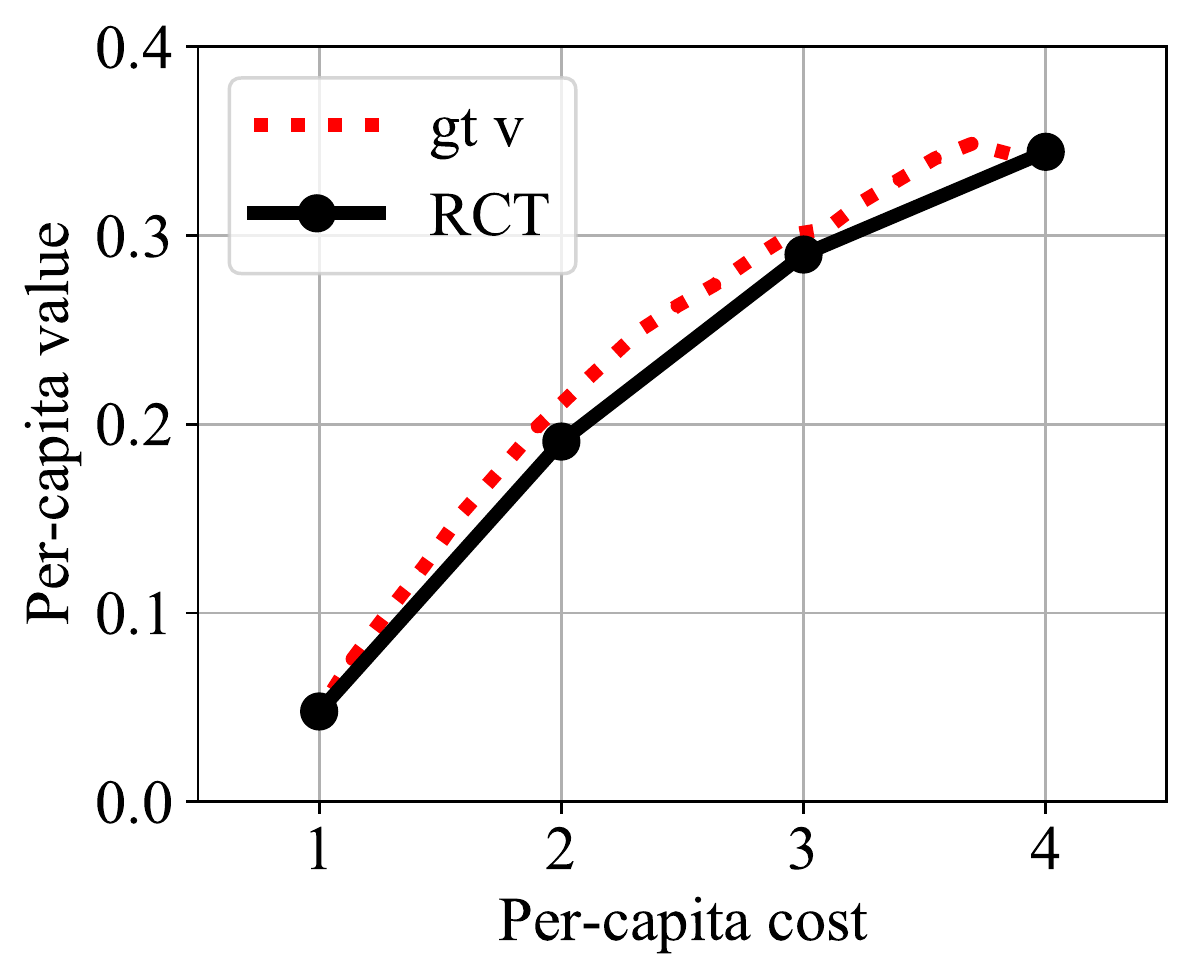}
   \caption{}\label{fig:synthetic_gt_rct}
  \end{subfigure}
  \begin{subfigure}[b]{0.48\linewidth}
   \includegraphics[width=1.0\linewidth]{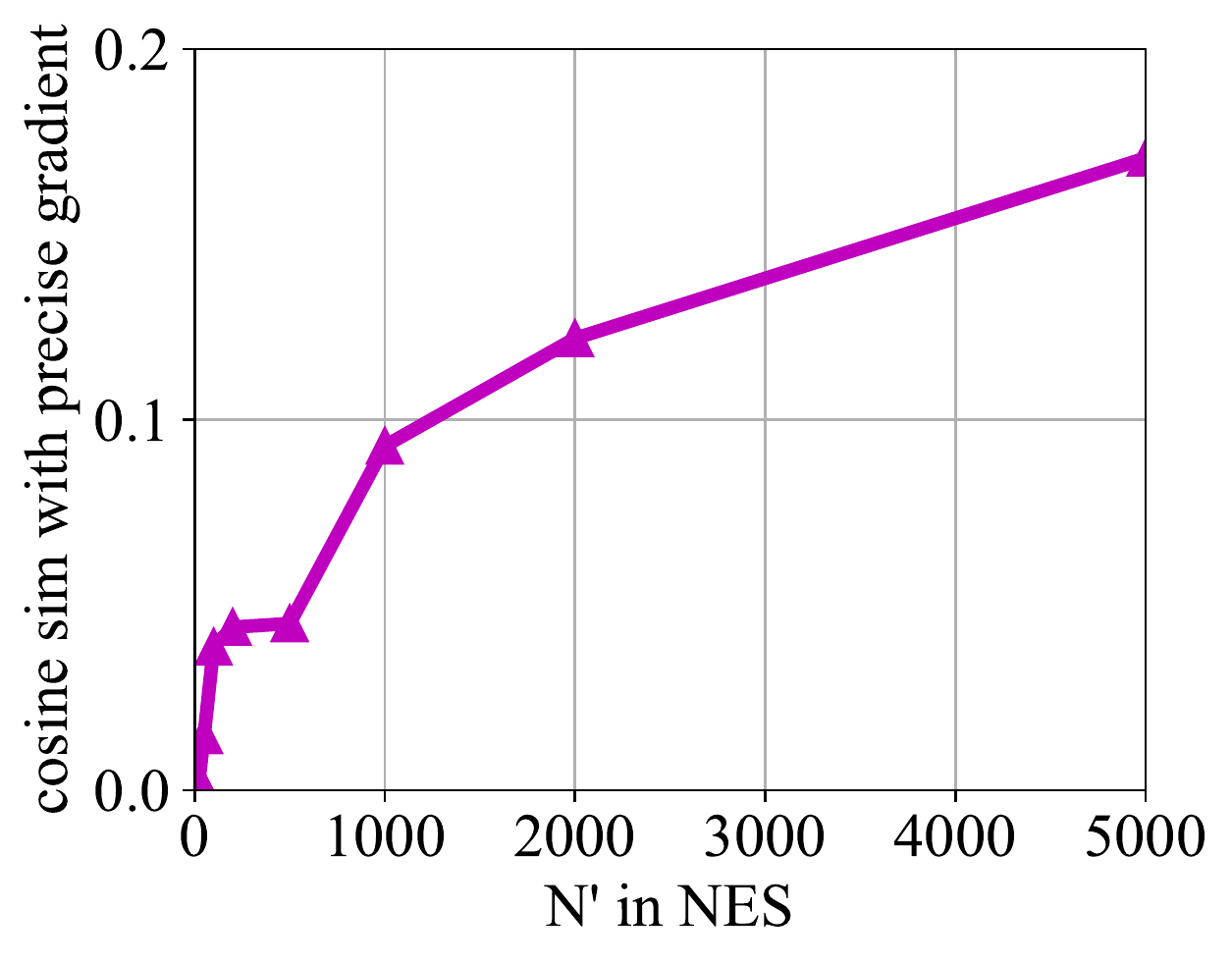}
   \caption{}\label{fig:synthetic_nes_cos_sim}
  \end{subfigure}

  \begin{subfigure}[b]{0.48\linewidth}
   \includegraphics[width=1.0\linewidth]{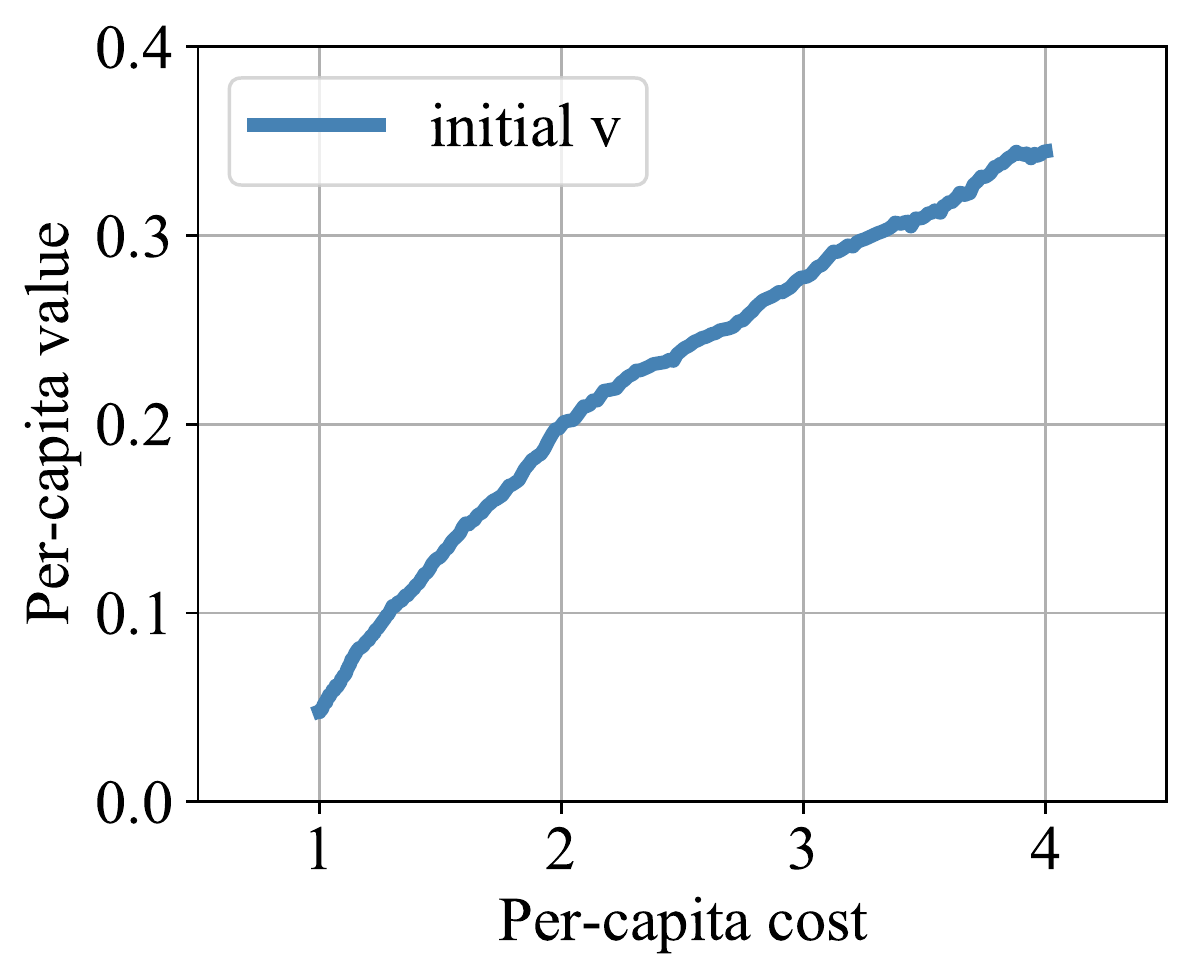}
   \caption{}\label{fig:synthetic_v0}
  \end{subfigure}
  \begin{subfigure}[b]{0.48\linewidth}
   \includegraphics[width=1.0\linewidth]{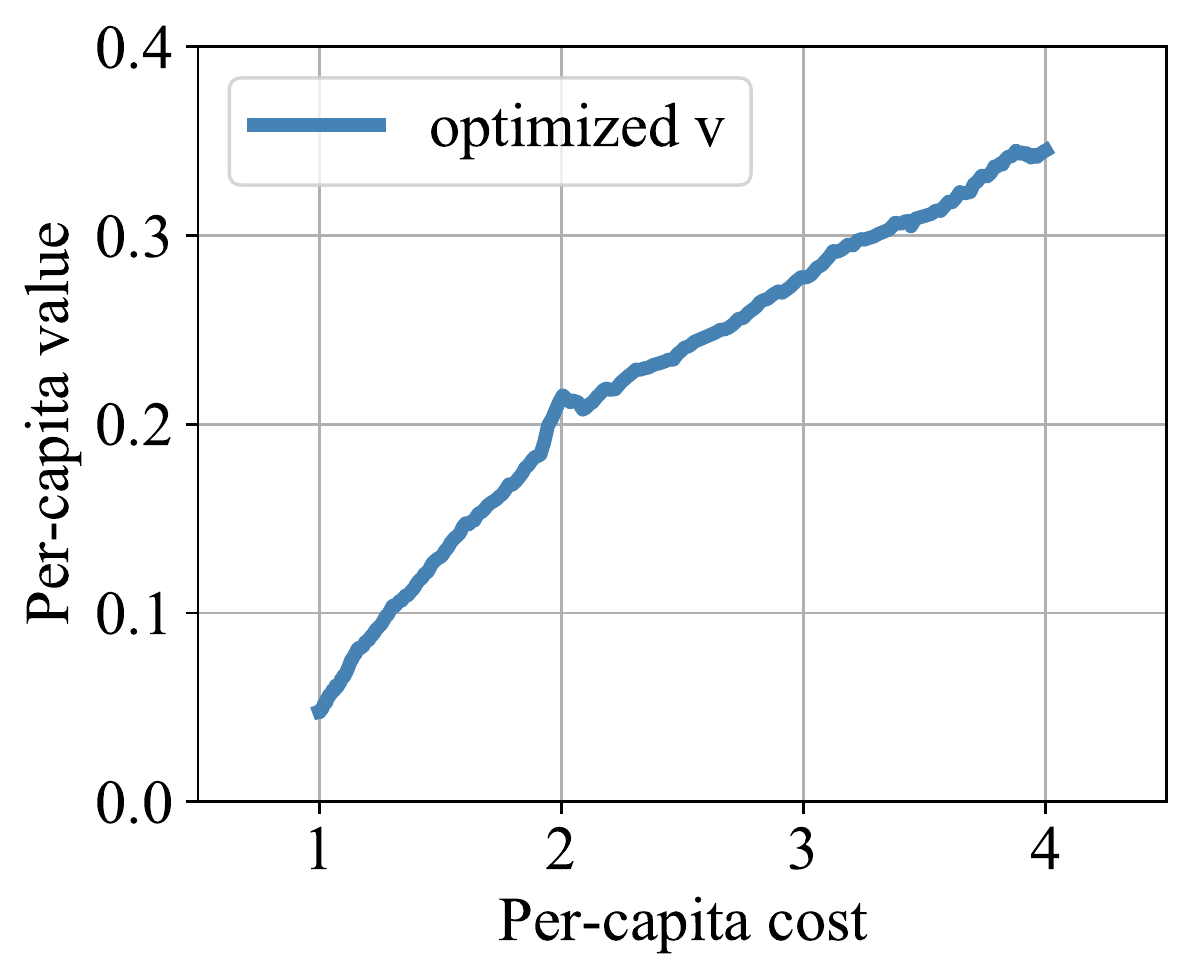}
   \caption{}\label{fig:synthetic_v100}
  \end{subfigure}
  \caption{Results on the synthetic dataset.}\vspace{-1.5em}
  \label{fig:synthetic_result}
 \end{figure}

\subsection{Results on Synthetic Dataset}
Results on the synthetic dataset are summarized in Figure~\ref{fig:synthetic_result}.
The black line in Figure~\ref{fig:synthetic_gt_rct} shows the statistics about per-capita cost and per-capita value in RCT.
As discussed in Section~\ref{sec:method_regularizer}, starting from any cost and value matrix we could obtain a series of budget allocation solutions corresponding to different budgets, and further invoke EOM to empirically estimate the per-capita value $V\left(v, c,\alpha\right)$ and per-capita cost $C\left(v, c,\alpha\right)$ for each budget allocation solution.
We perform the above EOM procedure using ground-truth $v^{\textrm{gt}}$ and $c^{\textrm{gt}}$, and plot the trajectory of $V\left(v^{\textrm{gt}}, c^{\textrm{gt}},\cdot\right)$ and $C\left(v^{\textrm{gt}}, c^{\textrm{gt}},\cdot\right)$ as the red dotted line in Figure~\ref{fig:synthetic_gt_rct}.
The value of $Q$ function for ground-truth $v^{\textrm{gt}}$ and $c^{\textrm{gt}}$ is $Q\left(v^{\textrm{gt}}, c^{\textrm{gt}}\right)=0.2123$.

To verify the effectiveness of the gradient estimation process, we first generate initial $v$ and $c$ as described in Section~\ref{sec:experiment_details} and study the cosine similarity between our estimated gradient and the precise gradient.
To this end, we first set $F'=40,000$ and use the finite-difference strategy to calculate the precise gradient $\frac{\partial Q(v,c)}{\partial v}$.
Then, we use NES to estimate $\frac{\partial Q(v,c)}{\partial v}$ (Equation~\eqref{eqn:nes2}) and gradually increase $N'$. 
Finally, we check the cosine similarity between the NES estimated gradient and the precise gradient.
Figure~\ref{fig:synthetic_nes_cos_sim} shows the relations between cosine similarity and $N'$, and we can see that as more sampling directions are involved in NES, the estimated gradients are more accurate.

We further plug the estimated gradient by NES into an Adam optimizer to maximize $Q(v, c)$, which represents the per-capita value when the per-capita cost equals 2.0.
Figure~\ref{fig:synthetic_v0} shows the per-capita value and per-capita cost of initial $v$ and $c$.
The value of $Q$ function for initial $v$ and $c$ is $Q(v, c)=0.1996$.
After 100 gradient ascent steps, we can increase the value of $Q$ function from 0.1996 to 0.2141.
Figure~\ref{fig:synthetic_v100} shows the per-capita value and per-capita cost of optimized $v$ and $c$.
By comparing Figure~\ref{fig:synthetic_v0} and Figure~\ref{fig:synthetic_v100}, we see that our proposed method could effectively increase the value of $Q$ function.
From the properties of EOM, we know that $Q$ would be an unbiased estimator of the marketing goal.
As such, the marketing goal could be effectively maximized.
Since we only have 10,000 data samples, the Adam optimizer could severely overfit the training data and thus in Figure~\ref{fig:synthetic_v100} there might be a peak around the region where the per-capita budget equals 2.0.

\subsection{Results on CRITEO-UPLIFT v2}
\begin{table}[thbp]
  \vspace{-1em}
  \caption{Results on CRITEO-UPLIFT v2.}
  \label{tab:criteo_result}
  \begin{tabular}{cc}
    \toprule
    Method & AUCC \\
    \midrule
    TSM-SL                           & 0.7561 $\pm$ 0.0113 \\
    Direct Rank~\cite{du2019improve} & 0.7562 $\pm$ 0.0131 \\   
    DRP~\cite{zhou2023direct}        & 0.7739 $\pm$ 0.0002 \\
    Ours-NES                         & 0.7849 $\pm$ 0.0167 \\
    Ours-FD                          & \bf{0.7904 $\pm$ 0.0096} \\
    \bottomrule
  \end{tabular}
\end{table}

Results on CRITEO-UPLIFT v2 are summarized in Table~\ref{tab:criteo_result}.
Ours-FD and Ours-NES refer to our proposed method with finite-difference strategy / NES as the gradient estimator, respectively.
We see both Ours-FD and Ours-NES significantly outperform all competitive methods in the sense of AUCC.
Ours-FD has the highest variance in AUCC for 20 runs, and this phenomenon could be partially explained by the fact that NES gradients are noisy.
Although the performance of Ours-FD surpasses that of Ours-NES, Ours-FD costs 4x more running times on an NVIDIA Tesla T4 GPU (Ours-FD 17.6 hours v.s. Ours-NES 4.3 hours) since we need to evaluate the $Q$ function for more times.

\subsection{Results on KUAISHOU-PRODUCE-COIN}
We adopt Ours-FD for the KUAISHOU-PRODUCE-COIN dataset and compare it with TSM-SL.

\begin{figure}[t]
  \centering
  \includegraphics[width=.75\linewidth]{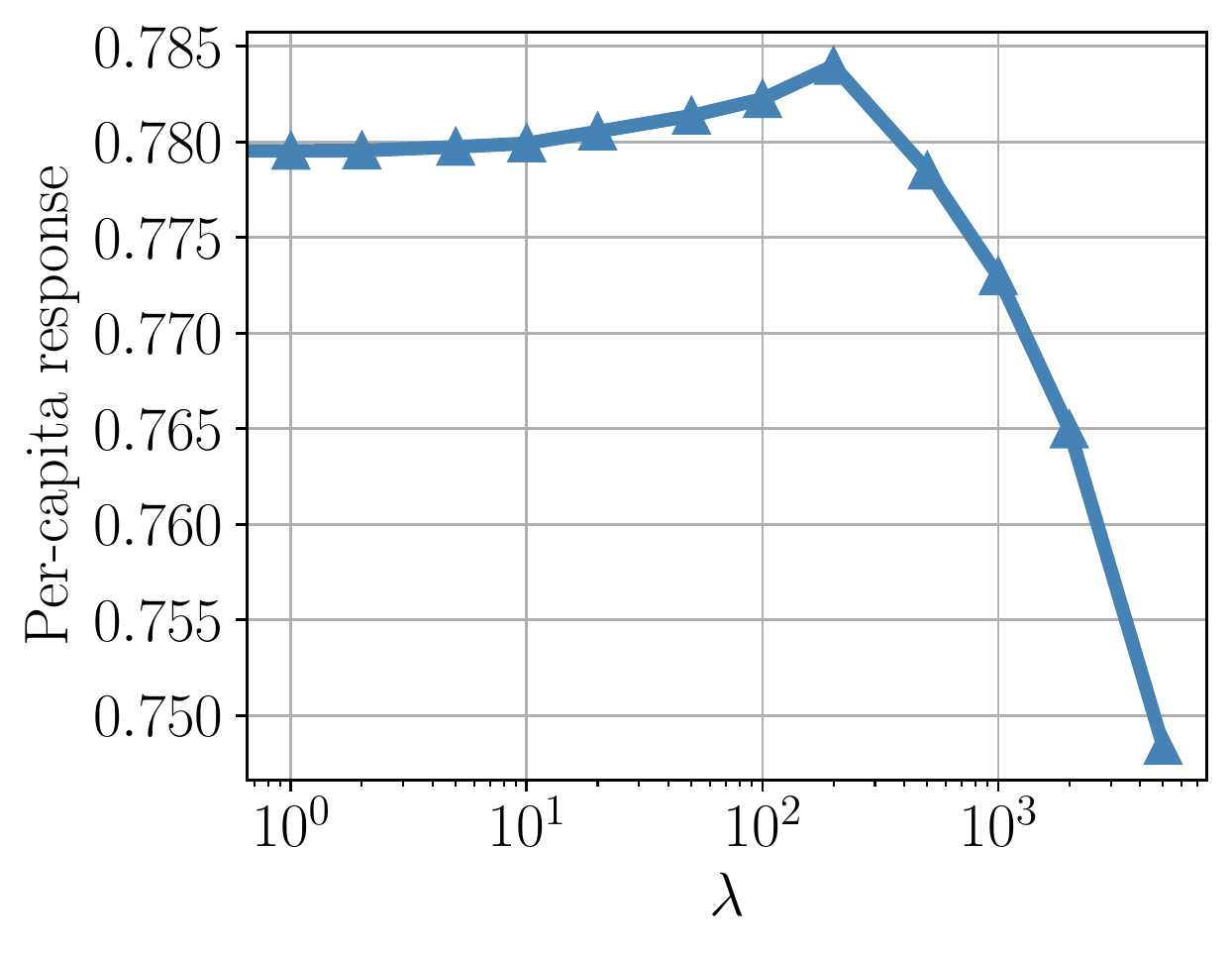}
  \caption{Impact of $\lambda$ for our method on the KUAISHOU-PRODUCE-COIN dataset.}
  \Description{Offline performance on KUAISHOU-PRODUCE-COIN}
  \label{fig:kuaishou_produce_coin_fd_lambda}
\end{figure}

\subsubsection{Offline Evaluation}
We are interested in how the value of $\lambda$ in Equation~\eqref{eqn:loss_surrogate} affects the performance of Ours-FD.
We set $\lambda\in\{0,1,2,5,10,20,50,100,200,500,1000,2000,5000\}$ and measure the performance of Ours-FD.
For each value of $\lambda$, we train the model four times and report their average performance.
Figure~\ref{fig:kuaishou_produce_coin_fd_lambda} illustrates the impact of $\lambda$ in Equation~\eqref{eqn:loss_surrogate} in Ours-FD.
By definition, we know that if we set $\lambda=0$, Ours-FD would be reduced to TSM-SL.
If $\lambda>0$, the regularizer would take effect and the DNN S-Learner model could learn from the marketing goal directly.
However, if $\lambda$ is too large, the regularizer could dominate the training objective and the predicted responses and costs could be far away from their ground-truth values thus the training process could be unstable.
In Figure~\ref{fig:kuaishou_produce_coin_fd_lambda}, when $\lambda\le 200$ the per-capita response would increase with $\lambda$, and it is clear that our regularizer is beneficial for the marketing goal in this region.
Furthermore, if $\lambda\ge500$, the performance of Ours-FD would be worse than that of TSM-SL since the training process becomes unstable.

\subsubsection{Online Evaluation}
For online evaluation, we re-train the model using the entire KUAISHOU-PRODUCE-COIN dataset.
We deploy Ours-FD with $\lambda=200$ and TSM-SL into the online A/B test.
To measure the performance of different methods, we calculate the per-capita response and per-capita cost by averaging the results over seven days.
When compared with TSM-SL, Ours-FD achieves a -0.369\% lower per-capita cost and a 1.24\% higher per-capita response simultaneously, which is considered to be a significant improvement over these months.
The Ours-FD model is currently deployed to allocate marketing budgets for hundreds of millions of users.

\section{Conclusion}\label{sec:conclusion}
In this paper, we investigate the end-to-end training framework in the budget allocation problem.
We formulate the business goal under a certain budget constraint as a black-box regularizer and develop two efficient gradient estimation algorithms to optimize it.
For both gradient estimation algorithms, we suggest a hyper-parameter to trade off the computation complexity and the accuracy of estimated gradients.
Our proposed method shall endow the well-trained DNN model with the ability to directly learn from the marketing goal and further maximize it.
Extensive experiments on three datasets have shown the superiority of our method in both offline simulation and online experiments.
Our future work will focus on reducing the variance for gradient estimation and applying this approach in more complicated real-world marketing scenarios.

%% The acknowledgments section is defined using the "acks" environment
%% (and NOT an unnumbered section). This ensures the proper
%% identification of the section in the article metadata, and the
%% consistent spelling of the heading.
%\begin{acks}
%To Haobin Lin and Shengjie Xue, for helpful discussions.
%\end{acks}

%%
%% The next two lines define the bibliography style to be used, and
%% the bibliography file.
%\newpage
%a
%\newpage
\bibliographystyle{ACM-Reference-Format}
\bibliography{coin_ref}

%%% -*-BibTeX-*-
%%% Do NOT edit. File created by BibTeX with style
%%% ACM-Reference-Format-Journals [18-Jan-2012].

\begin{thebibliography}{42}

%%% ====================================================================
%%% NOTE TO THE USER: you can override these defaults by providing
%%% customized versions of any of these macros before the \bibliography
%%% command.  Each of them MUST provide its own final punctuation,
%%% except for \shownote{}, \showDOI{}, and \showURL{}.  The latter two
%%% do not use final punctuation, in order to avoid confusing it with
%%% the Web address.
%%%
%%% To suppress output of a particular field, define its macro to expand
%%% to an empty string, or better, \unskip, like this:
%%%
%%% \newcommand{\showDOI}[1]{\unskip}   % LaTeX syntax
%%%
%%% \def \showDOI #1{\unskip}           % plain TeX syntax
%%%
%%% ====================================================================

\ifx \showCODEN    \undefined \def \showCODEN     #1{\unskip}     \fi
\ifx \showDOI      \undefined \def \showDOI       #1{#1}\fi
\ifx \showISBNx    \undefined \def \showISBNx     #1{\unskip}     \fi
\ifx \showISBNxiii \undefined \def \showISBNxiii  #1{\unskip}     \fi
\ifx \showISSN     \undefined \def \showISSN      #1{\unskip}     \fi
\ifx \showLCCN     \undefined \def \showLCCN      #1{\unskip}     \fi
\ifx \shownote     \undefined \def \shownote      #1{#1}          \fi
\ifx \showarticletitle \undefined \def \showarticletitle #1{#1}   \fi
\ifx \showURL      \undefined \def \showURL       {\relax}        \fi
% The following commands are used for tagged output and should be
% invisible to TeX
\providecommand\bibfield[2]{#2}
\providecommand\bibinfo[2]{#2}
\providecommand\natexlab[1]{#1}
\providecommand\showeprint[2][]{arXiv:#2}

\bibitem[Abadi et~al\mbox{.}(2015)]%
        {tensorflow2015-whitepaper}
\bibfield{author}{\bibinfo{person}{Mart\'{i}n Abadi}, \bibinfo{person}{Ashish
  Agarwal}, \bibinfo{person}{Paul Barham}, \bibinfo{person}{Eugene Brevdo},
  \bibinfo{person}{Zhifeng Chen}, \bibinfo{person}{Craig Citro},
  \bibinfo{person}{Greg~S. Corrado}, \bibinfo{person}{Andy Davis},
  \bibinfo{person}{Jeffrey Dean}, \bibinfo{person}{Matthieu Devin},
  \bibinfo{person}{Sanjay Ghemawat}, \bibinfo{person}{Ian Goodfellow},
  \bibinfo{person}{Andrew Harp}, \bibinfo{person}{Geoffrey Irving},
  \bibinfo{person}{Michael Isard}, \bibinfo{person}{Yangqing Jia},
  \bibinfo{person}{Rafal Jozefowicz}, \bibinfo{person}{Lukasz Kaiser},
  \bibinfo{person}{Manjunath Kudlur}, \bibinfo{person}{Josh Levenberg},
  \bibinfo{person}{Dandelion Man\'{e}}, \bibinfo{person}{Rajat Monga},
  \bibinfo{person}{Sherry Moore}, \bibinfo{person}{Derek Murray},
  \bibinfo{person}{Chris Olah}, \bibinfo{person}{Mike Schuster},
  \bibinfo{person}{Jonathon Shlens}, \bibinfo{person}{Benoit Steiner},
  \bibinfo{person}{Ilya Sutskever}, \bibinfo{person}{Kunal Talwar},
  \bibinfo{person}{Paul Tucker}, \bibinfo{person}{Vincent Vanhoucke},
  \bibinfo{person}{Vijay Vasudevan}, \bibinfo{person}{Fernanda Vi\'{e}gas},
  \bibinfo{person}{Oriol Vinyals}, \bibinfo{person}{Pete Warden},
  \bibinfo{person}{Martin Wattenberg}, \bibinfo{person}{Martin Wicke},
  \bibinfo{person}{Yuan Yu}, {and} \bibinfo{person}{Xiaoqiang Zheng}.}
  \bibinfo{year}{2015}\natexlab{}.
\newblock \bibinfo{title}{{TensorFlow}: Large-Scale Machine Learning on
  Heterogeneous Systems}.
\newblock
\newblock
\urldef\tempurl%
\url{https://www.tensorflow.org/}
\showURL{%
\tempurl}
\newblock
\shownote{Software available from tensorflow.org}.


\bibitem[Ai et~al\mbox{.}(2022)]%
        {ai2022lbcf}
\bibfield{author}{\bibinfo{person}{Meng Ai}, \bibinfo{person}{Biao Li},
  \bibinfo{person}{Heyang Gong}, \bibinfo{person}{Qingwei Yu},
  \bibinfo{person}{Shengjie Xue}, \bibinfo{person}{Yuan Zhang},
  \bibinfo{person}{Yunzhou Zhang}, {and} \bibinfo{person}{Peng Jiang}.}
  \bibinfo{year}{2022}\natexlab{}.
\newblock \showarticletitle{LBCF: A Large-Scale Budget-Constrained Causal
  Forest Algorithm}. In \bibinfo{booktitle}{\emph{WWW 2022}}.
  \bibinfo{pages}{2310--2319}.
\newblock


\bibitem[Albert and Goldenberg(2022)]%
        {albert2022commerce}
\bibfield{author}{\bibinfo{person}{Javier Albert} {and} \bibinfo{person}{Dmitri
  Goldenberg}.} \bibinfo{year}{2022}\natexlab{}.
\newblock \showarticletitle{E-Commerce Promotions Personalization via Online
  Multiple-Choice Knapsack with Uplift Modeling}. In
  \bibinfo{booktitle}{\emph{CIKM 2022}}. \bibinfo{pages}{2863--2872}.
\newblock


\bibitem[Athey and Imbens(2016)]%
        {athey2016recursive}
\bibfield{author}{\bibinfo{person}{Susan Athey} {and} \bibinfo{person}{Guido
  Imbens}.} \bibinfo{year}{2016}\natexlab{}.
\newblock \showarticletitle{Recursive partitioning for heterogeneous causal
  effects}.
\newblock \bibinfo{journal}{\emph{Proceedings of the National Academy of
  Sciences}} \bibinfo{volume}{113}, \bibinfo{number}{27}
  (\bibinfo{year}{2016}), \bibinfo{pages}{7353--7360}.
\newblock


\bibitem[Athey et~al\mbox{.}(2019)]%
        {athey2019generalized}
\bibfield{author}{\bibinfo{person}{Susan Athey}, \bibinfo{person}{Julie
  Tibshirani}, {and} \bibinfo{person}{Stefan Wager}.}
  \bibinfo{year}{2019}\natexlab{}.
\newblock \showarticletitle{Generalized random forests}.
\newblock \bibinfo{journal}{\emph{The Annals of Statistics}}
  \bibinfo{volume}{47}, \bibinfo{number}{2} (\bibinfo{year}{2019}),
  \bibinfo{pages}{1148--1178}.
\newblock


\bibitem[Chen et~al\mbox{.}(2017)]%
        {chen2017zoo}
\bibfield{author}{\bibinfo{person}{Pin-Yu Chen}, \bibinfo{person}{Huan Zhang},
  \bibinfo{person}{Yash Sharma}, \bibinfo{person}{Jinfeng Yi}, {and}
  \bibinfo{person}{Cho-Jui Hsieh}.} \bibinfo{year}{2017}\natexlab{}.
\newblock \showarticletitle{Zoo: Zeroth order optimization based black-box
  attacks to deep neural networks without training substitute models}. In
  \bibinfo{booktitle}{\emph{Proceedings of the 10th ACM workshop on artificial
  intelligence and security}}. \bibinfo{pages}{15--26}.
\newblock


\bibitem[Diemert et~al\mbox{.}(2018)]%
        {diemert2018large}
\bibfield{author}{\bibinfo{person}{Eustache Diemert}, \bibinfo{person}{Artem
  Betlei}, \bibinfo{person}{Christophe Renaudin}, {and}
  \bibinfo{person}{Massih-Reza Amini}.} \bibinfo{year}{2018}\natexlab{}.
\newblock \showarticletitle{A large scale benchmark for uplift modeling}. In
  \bibinfo{booktitle}{\emph{SIGKDD 2018}}.
\newblock


\bibitem[Du et~al\mbox{.}(2019)]%
        {du2019improve}
\bibfield{author}{\bibinfo{person}{Shuyang Du}, \bibinfo{person}{James Lee},
  {and} \bibinfo{person}{Farzin Ghaffarizadeh}.}
  \bibinfo{year}{2019}\natexlab{}.
\newblock \showarticletitle{Improve User Retention with Causal Learning}. In
  \bibinfo{booktitle}{\emph{SIGKDD Workshop on Causal Discovery 2019}}. PMLR,
  \bibinfo{pages}{34--49}.
\newblock


\bibitem[Goldenberg et~al\mbox{.}(2020)]%
        {goldenberg2020free}
\bibfield{author}{\bibinfo{person}{Dmitri Goldenberg}, \bibinfo{person}{Javier
  Albert}, \bibinfo{person}{Lucas Bernardi}, {and} \bibinfo{person}{Pablo
  Estevez}.} \bibinfo{year}{2020}\natexlab{}.
\newblock \showarticletitle{Free lunch! retrospective uplift modeling for
  dynamic promotions recommendation within roi constraints}. In
  \bibinfo{booktitle}{\emph{ACM Conference on Recommender Systems 2020}}.
  \bibinfo{pages}{486--491}.
\newblock


\bibitem[Guelman et~al\mbox{.}(2015)]%
        {guelman2015uplift}
\bibfield{author}{\bibinfo{person}{Leo Guelman}, \bibinfo{person}{Montserrat
  Guill{\'e}n}, {and} \bibinfo{person}{Ana~M P{\'e}rez-Mar{\'\i}n}.}
  \bibinfo{year}{2015}\natexlab{}.
\newblock \showarticletitle{Uplift random forests}.
\newblock \bibinfo{journal}{\emph{Cybernetics and Systems}}
  \bibinfo{volume}{46}, \bibinfo{number}{3-4} (\bibinfo{year}{2015}),
  \bibinfo{pages}{230--248}.
\newblock


\bibitem[Guo et~al\mbox{.}(2019)]%
        {guo2019subspace}
\bibfield{author}{\bibinfo{person}{Yiwen Guo}, \bibinfo{person}{Ziang Yan},
  {and} \bibinfo{person}{Changshui Zhang}.} \bibinfo{year}{2019}\natexlab{}.
\newblock \showarticletitle{Subspace attack: Exploiting promising subspaces for
  query-efficient black-box attacks}.
\newblock \bibinfo{journal}{\emph{NeurIPS 2019}}  \bibinfo{volume}{32}.
\newblock


\bibitem[He et~al\mbox{.}(2015)]%
        {he2015delving}
\bibfield{author}{\bibinfo{person}{Kaiming He}, \bibinfo{person}{Xiangyu
  Zhang}, \bibinfo{person}{Shaoqing Ren}, {and} \bibinfo{person}{Jian Sun}.}
  \bibinfo{year}{2015}\natexlab{}.
\newblock \showarticletitle{Delving deep into rectifiers: Surpassing
  human-level performance on imagenet classification}. In
  \bibinfo{booktitle}{\emph{ICCV 2015}}. \bibinfo{pages}{1026--1034}.
\newblock


\bibitem[Ilyas et~al\mbox{.}(2018)]%
        {ilyas2018black}
\bibfield{author}{\bibinfo{person}{Andrew Ilyas}, \bibinfo{person}{Logan
  Engstrom}, \bibinfo{person}{Anish Athalye}, {and} \bibinfo{person}{Jessy
  Lin}.} \bibinfo{year}{2018}\natexlab{}.
\newblock \showarticletitle{Black-box adversarial attacks with limited queries
  and information}. In \bibinfo{booktitle}{\emph{ICML 2018}}. PMLR,
  \bibinfo{pages}{2137--2146}.
\newblock


\bibitem[Ilyas et~al\mbox{.}(2019)]%
        {ilyas2018prior}
\bibfield{author}{\bibinfo{person}{Andrew Ilyas}, \bibinfo{person}{Logan
  Engstrom}, {and} \bibinfo{person}{Aleksander Madry}.}
  \bibinfo{year}{2019}\natexlab{}.
\newblock \showarticletitle{Prior Convictions: Black-box Adversarial Attacks
  with Bandits and Priors}. In \bibinfo{booktitle}{\emph{ICLR 2019}}.
\newblock


\bibitem[Ioffe and Szegedy(2015)]%
        {ioffe2015batch}
\bibfield{author}{\bibinfo{person}{Sergey Ioffe} {and}
  \bibinfo{person}{Christian Szegedy}.} \bibinfo{year}{2015}\natexlab{}.
\newblock \showarticletitle{Batch normalization: Accelerating deep network
  training by reducing internal covariate shift}. In
  \bibinfo{booktitle}{\emph{ICML 2015}}. pmlr, \bibinfo{pages}{448--456}.
\newblock


\bibitem[Kellerer et~al\mbox{.}(2004)]%
        {kellerer2004multiple}
\bibfield{author}{\bibinfo{person}{Hans Kellerer}, \bibinfo{person}{Ulrich
  Pferschy}, {and} \bibinfo{person}{David Pisinger}.}
  \bibinfo{year}{2004}\natexlab{}.
\newblock \showarticletitle{The multiple-choice knapsack problem}.
\newblock In \bibinfo{booktitle}{\emph{Knapsack Problems}}.
  \bibinfo{publisher}{Springer}, \bibinfo{pages}{317--347}.
\newblock


\bibitem[Kingma and Ba(2015)]%
        {kingma2015adam}
\bibfield{author}{\bibinfo{person}{Diederik~P Kingma} {and}
  \bibinfo{person}{Jimmy Ba}.} \bibinfo{year}{2015}\natexlab{}.
\newblock \showarticletitle{Adam: A Method for Stochastic Optimization}. In
  \bibinfo{booktitle}{\emph{ICLR 2015}}.
\newblock


\bibitem[Krizhevsky et~al\mbox{.}(2012)]%
        {krizhevsky2012imagenet}
\bibfield{author}{\bibinfo{person}{Alex Krizhevsky}, \bibinfo{person}{Ilya
  Sutskever}, {and} \bibinfo{person}{Geoffrey~E Hinton}.}
  \bibinfo{year}{2012}\natexlab{}.
\newblock \showarticletitle{Imagenet classification with deep convolutional
  networks}. In \bibinfo{booktitle}{\emph{NIPS 2012}},
  Vol.~\bibinfo{volume}{1097}.
\newblock


\bibitem[K{\"u}nzel et~al\mbox{.}(2019)]%
        {kunzel2019metalearners}
\bibfield{author}{\bibinfo{person}{S{\"o}ren~R K{\"u}nzel},
  \bibinfo{person}{Jasjeet~S Sekhon}, \bibinfo{person}{Peter~J Bickel}, {and}
  \bibinfo{person}{Bin Yu}.} \bibinfo{year}{2019}\natexlab{}.
\newblock \showarticletitle{Metalearners for estimating heterogeneous treatment
  effects using machine learning}.
\newblock \bibinfo{journal}{\emph{Proceedings of the national academy of
  sciences}} \bibinfo{volume}{116}, \bibinfo{number}{10}
  (\bibinfo{year}{2019}), \bibinfo{pages}{4156--4165}.
\newblock


\bibitem[Lin et~al\mbox{.}(2017)]%
        {lin2017monetary}
\bibfield{author}{\bibinfo{person}{Ying-Chun Lin}, \bibinfo{person}{Chi-Hsuan
  Huang}, \bibinfo{person}{Chu-Cheng Hsieh}, \bibinfo{person}{Yu-Chen Shu},
  {and} \bibinfo{person}{Kun-Ta Chuang}.} \bibinfo{year}{2017}\natexlab{}.
\newblock \showarticletitle{Monetary discount strategies for real-time
  promotion campaign}. In \bibinfo{booktitle}{\emph{WWW 2017}}.
  \bibinfo{pages}{1123--1132}.
\newblock


\bibitem[Makhijani et~al\mbox{.}(2019)]%
        {makhijani2019lore}
\bibfield{author}{\bibinfo{person}{Rahul Makhijani}, \bibinfo{person}{Shreya
  Chakrabarti}, \bibinfo{person}{Dale Struble}, {and} \bibinfo{person}{Yi
  Liu}.} \bibinfo{year}{2019}\natexlab{}.
\newblock \showarticletitle{LORE: a large-scale offer recommendation engine
  with eligibility and capacity constraints}. In \bibinfo{booktitle}{\emph{ACM
  Conference on Recommender Systems 2019}}. \bibinfo{pages}{160--168}.
\newblock


\bibitem[Nie and Wager(2021)]%
        {nie2021quasi}
\bibfield{author}{\bibinfo{person}{Xinkun Nie} {and} \bibinfo{person}{Stefan
  Wager}.} \bibinfo{year}{2021}\natexlab{}.
\newblock \showarticletitle{Quasi-oracle estimation of heterogeneous treatment
  effects}.
\newblock \bibinfo{journal}{\emph{Biometrika}} \bibinfo{volume}{108},
  \bibinfo{number}{2} (\bibinfo{year}{2021}), \bibinfo{pages}{299--319}.
\newblock


\bibitem[Oprescu et~al\mbox{.}(2019)]%
        {oprescu2019orthogonal}
\bibfield{author}{\bibinfo{person}{Miruna Oprescu}, \bibinfo{person}{Vasilis
  Syrgkanis}, {and} \bibinfo{person}{Zhiwei~Steven Wu}.}
  \bibinfo{year}{2019}\natexlab{}.
\newblock \showarticletitle{Orthogonal random forest for causal inference}. In
  \bibinfo{booktitle}{\emph{ICML 2019}}. PMLR, \bibinfo{pages}{4932--4941}.
\newblock


\bibitem[Paszke et~al\mbox{.}(2019)]%
        {paszke2019pytorch}
\bibfield{author}{\bibinfo{person}{Adam Paszke}, \bibinfo{person}{Sam Gross},
  \bibinfo{person}{Francisco Massa}, \bibinfo{person}{Adam Lerer},
  \bibinfo{person}{James Bradbury}, \bibinfo{person}{Gregory Chanan},
  \bibinfo{person}{Trevor Killeen}, \bibinfo{person}{Zeming Lin},
  \bibinfo{person}{Natalia Gimelshein}, \bibinfo{person}{Luca Antiga},
  {et~al\mbox{.}}} \bibinfo{year}{2019}\natexlab{}.
\newblock \showarticletitle{Pytorch: An imperative style, high-performance deep
  learning library}. In \bibinfo{booktitle}{\emph{NeurIPS 2019}},
  Vol.~\bibinfo{volume}{32}.
\newblock


\bibitem[Peter and Maria(2014)]%
        {peter2014calculus}
\bibfield{author}{\bibinfo{person}{D~Lax Peter} {and}
  \bibinfo{person}{Shea~Terrell Maria}.} \bibinfo{year}{2014}\natexlab{}.
\newblock \bibinfo{title}{Calculus With Applications}.
\newblock
\newblock


\bibitem[Phillips(2021)]%
        {phillips2021pricing}
\bibfield{author}{\bibinfo{person}{Robert~L Phillips}.}
  \bibinfo{year}{2021}\natexlab{}.
\newblock \showarticletitle{Pricing and revenue optimization}.
\newblock In \bibinfo{booktitle}{\emph{Pricing and Revenue Optimization}}.
  \bibinfo{publisher}{Stanford university press}.
\newblock


\bibitem[Rzepakowski and Jaroszewicz(2012)]%
        {rzepakowski2012decision}
\bibfield{author}{\bibinfo{person}{Piotr Rzepakowski} {and}
  \bibinfo{person}{Szymon Jaroszewicz}.} \bibinfo{year}{2012}\natexlab{}.
\newblock \showarticletitle{Decision trees for uplift modeling with single and
  multiple treatments}.
\newblock \bibinfo{journal}{\emph{Knowledge and Information Systems}}
  \bibinfo{volume}{32}, \bibinfo{number}{2} (\bibinfo{year}{2012}),
  \bibinfo{pages}{303--327}.
\newblock


\bibitem[Salimans et~al\mbox{.}(2017)]%
        {salimans2017evolution}
\bibfield{author}{\bibinfo{person}{Tim Salimans}, \bibinfo{person}{Jonathan
  Ho}, \bibinfo{person}{Xi Chen}, \bibinfo{person}{Szymon Sidor}, {and}
  \bibinfo{person}{Ilya Sutskever}.} \bibinfo{year}{2017}\natexlab{}.
\newblock \showarticletitle{Evolution strategies as a scalable alternative to
  reinforcement learning}.
\newblock \bibinfo{journal}{\emph{arXiv preprint arXiv:1703.03864}}
  (\bibinfo{year}{2017}).
\newblock


\bibitem[So{\l}tys et~al\mbox{.}(2015)]%
        {soltys2015ensemble}
\bibfield{author}{\bibinfo{person}{Micha{\l} So{\l}tys},
  \bibinfo{person}{Szymon Jaroszewicz}, {and} \bibinfo{person}{Piotr
  Rzepakowski}.} \bibinfo{year}{2015}\natexlab{}.
\newblock \showarticletitle{Ensemble methods for uplift modeling}.
\newblock \bibinfo{journal}{\emph{Data Mining and Knowledge Discovery}}
  \bibinfo{volume}{6}, \bibinfo{number}{29} (\bibinfo{year}{2015}),
  \bibinfo{pages}{1531--1559}.
\newblock


\bibitem[Sutton et~al\mbox{.}(1999)]%
        {sutton1999policy}
\bibfield{author}{\bibinfo{person}{Richard~S Sutton}, \bibinfo{person}{David
  McAllester}, \bibinfo{person}{Satinder Singh}, {and} \bibinfo{person}{Yishay
  Mansour}.} \bibinfo{year}{1999}\natexlab{}.
\newblock \showarticletitle{Policy gradient methods for reinforcement learning
  with function approximation}.
\newblock \bibinfo{journal}{\emph{NeurIPS 1999}}  \bibinfo{volume}{12}
  (\bibinfo{year}{1999}).
\newblock


\bibitem[Tu et~al\mbox{.}(2019)]%
        {tu2019autozoom}
\bibfield{author}{\bibinfo{person}{Chun-Chen Tu}, \bibinfo{person}{Paishun
  Ting}, \bibinfo{person}{Pin-Yu Chen}, \bibinfo{person}{Sijia Liu},
  \bibinfo{person}{Huan Zhang}, \bibinfo{person}{Jinfeng Yi},
  \bibinfo{person}{Cho-Jui Hsieh}, {and} \bibinfo{person}{Shin-Ming Cheng}.}
  \bibinfo{year}{2019}\natexlab{}.
\newblock \showarticletitle{Autozoom: Autoencoder-based zeroth order
  optimization method for attacking black-box neural networks}. In
  \bibinfo{booktitle}{\emph{AAAI 2019}}, Vol.~\bibinfo{volume}{33}.
  \bibinfo{pages}{742--749}.
\newblock


\bibitem[Tu et~al\mbox{.}(2021)]%
        {tu2021personalized}
\bibfield{author}{\bibinfo{person}{Ye Tu}, \bibinfo{person}{Kinjal Basu},
  \bibinfo{person}{Cyrus DiCiccio}, \bibinfo{person}{Romil Bansal},
  \bibinfo{person}{Preetam Nandy}, \bibinfo{person}{Padmini Jaikumar}, {and}
  \bibinfo{person}{Shaunak Chatterjee}.} \bibinfo{year}{2021}\natexlab{}.
\newblock \showarticletitle{Personalized treatment selection using causal
  heterogeneity}. In \bibinfo{booktitle}{\emph{WWW 2021}}.
  \bibinfo{pages}{1574--1585}.
\newblock


\bibitem[Wager and Athey(2018)]%
        {wager2018estimation}
\bibfield{author}{\bibinfo{person}{Stefan Wager} {and} \bibinfo{person}{Susan
  Athey}.} \bibinfo{year}{2018}\natexlab{}.
\newblock \showarticletitle{Estimation and inference of heterogeneous treatment
  effects using random forests}.
\newblock \bibinfo{journal}{\emph{J. Amer. Statist. Assoc.}}
  \bibinfo{volume}{113}, \bibinfo{number}{523} (\bibinfo{year}{2018}),
  \bibinfo{pages}{1228--1242}.
\newblock


\bibitem[Wierstra et~al\mbox{.}(2014)]%
        {wierstra2014natural}
\bibfield{author}{\bibinfo{person}{Daan Wierstra}, \bibinfo{person}{Tom
  Schaul}, \bibinfo{person}{Tobias Glasmachers}, \bibinfo{person}{Yi Sun},
  \bibinfo{person}{Jan Peters}, {and} \bibinfo{person}{J{\"u}rgen
  Schmidhuber}.} \bibinfo{year}{2014}\natexlab{}.
\newblock \showarticletitle{Natural evolution strategies}.
\newblock \bibinfo{journal}{\emph{JMLR 2014}} \bibinfo{volume}{15},
  \bibinfo{number}{1} (\bibinfo{year}{2014}), \bibinfo{pages}{949--980}.
\newblock


\bibitem[Xiao et~al\mbox{.}(2019)]%
        {xiao2019model}
\bibfield{author}{\bibinfo{person}{Shuai Xiao}, \bibinfo{person}{Le Guo},
  \bibinfo{person}{Zaifan Jiang}, \bibinfo{person}{Lei Lv},
  \bibinfo{person}{Yuanbo Chen}, \bibinfo{person}{Jun Zhu}, {and}
  \bibinfo{person}{Shuang Yang}.} \bibinfo{year}{2019}\natexlab{}.
\newblock \showarticletitle{Model-based constrained MDP for budget allocation
  in sequential incentive marketing}. In \bibinfo{booktitle}{\emph{CIKM 2019}}.
  \bibinfo{pages}{971--980}.
\newblock


\bibitem[Zemel(1984)]%
        {zemel1984n}
\bibfield{author}{\bibinfo{person}{Eitan Zemel}.}
  \bibinfo{year}{1984}\natexlab{}.
\newblock \showarticletitle{An O (n) algorithm for the linear multiple choice
  knapsack problem and related problems}.
\newblock \bibinfo{journal}{\emph{Information processing letters}}
  \bibinfo{volume}{18}, \bibinfo{number}{3} (\bibinfo{year}{1984}),
  \bibinfo{pages}{123--128}.
\newblock


\bibitem[Zhang et~al\mbox{.}(2021)]%
        {zhang2021bcorle}
\bibfield{author}{\bibinfo{person}{Yang Zhang}, \bibinfo{person}{Bo Tang},
  \bibinfo{person}{Qingyu Yang}, \bibinfo{person}{Dou An},
  \bibinfo{person}{Hongyin Tang}, \bibinfo{person}{Chenyang Xi},
  \bibinfo{person}{Xueying Li}, {and} \bibinfo{person}{Feiyu Xiong}.}
  \bibinfo{year}{2021}\natexlab{}.
\newblock \showarticletitle{BCORLE ($\lambda$): An Offline Reinforcement
  Learning and Evaluation Framework for Coupons Allocation in E-commerce
  Market}. In \bibinfo{booktitle}{\emph{NeurIPS 2021}},
  Vol.~\bibinfo{volume}{34}. \bibinfo{pages}{20410--20422}.
\newblock


\bibitem[Zhao et~al\mbox{.}(2019)]%
        {zhao2019unified}
\bibfield{author}{\bibinfo{person}{Kui Zhao}, \bibinfo{person}{Junhao Hua},
  \bibinfo{person}{Ling Yan}, \bibinfo{person}{Qi Zhang}, \bibinfo{person}{Huan
  Xu}, {and} \bibinfo{person}{Cheng Yang}.} \bibinfo{year}{2019}\natexlab{}.
\newblock \showarticletitle{A Unified Framework for Marketing Budget
  Allocation}. In \bibinfo{booktitle}{\emph{SIGKDD 2019}}.
  \bibinfo{pages}{1820--1830}.
\newblock


\bibitem[Zhao et~al\mbox{.}(2017)]%
        {zhao2017uplift}
\bibfield{author}{\bibinfo{person}{Yan Zhao}, \bibinfo{person}{Xiao Fang},
  {and} \bibinfo{person}{David Simchi-Levi}.} \bibinfo{year}{2017}\natexlab{}.
\newblock \showarticletitle{Uplift modeling with multiple treatments and
  general response types}. In \bibinfo{booktitle}{\emph{SIAM 2017}}. SIAM,
  \bibinfo{pages}{588--596}.
\newblock


\bibitem[Zhao and Harinen(2019)]%
        {zhao2019uplift}
\bibfield{author}{\bibinfo{person}{Zhenyu Zhao} {and} \bibinfo{person}{Totte
  Harinen}.} \bibinfo{year}{2019}\natexlab{}.
\newblock \showarticletitle{Uplift modeling for multiple treatments with cost
  optimization}. In \bibinfo{booktitle}{\emph{IEEE International Conference on
  Data Science and Advanced Analytics (DSAA) 2019}}. IEEE,
  \bibinfo{pages}{422--431}.
\newblock


\bibitem[Zhou et~al\mbox{.}(2023)]%
        {zhou2023direct}
\bibfield{author}{\bibinfo{person}{Hao Zhou}, \bibinfo{person}{Shaoming Li},
  \bibinfo{person}{Guibin Jiang}, \bibinfo{person}{Jiaqi Zheng}, {and}
  \bibinfo{person}{Dong Wang}.} \bibinfo{year}{2023}\natexlab{}.
\newblock \showarticletitle{Direct Heterogeneous Causal Learning for Resource
  Allocation Problems in Marketing}. In \bibinfo{booktitle}{\emph{AAAI 2023}}.
\newblock


\bibitem[Zou et~al\mbox{.}(2020)]%
        {zou2020heterogeneous}
\bibfield{author}{\bibinfo{person}{Will~Y Zou}, \bibinfo{person}{Shuyang Du},
  \bibinfo{person}{James Lee}, {and} \bibinfo{person}{Jan Pedersen}.}
  \bibinfo{year}{2020}\natexlab{}.
\newblock \showarticletitle{Heterogeneous causal learning for effectiveness
  optimization in user marketing}.
\newblock \bibinfo{journal}{\emph{arXiv preprint arXiv:2004.09702}}
  (\bibinfo{year}{2020}).
\newblock


\end{thebibliography}

%%
%% If your work has an appendix, this is the place to put it.
%\appendix

%\section{Derivation of MCKP's Dual Problem}\label{sec:appendix_mckp_dual}
%In this section we derive the bi-section dual problem method for effectively solving MCKPs from the first principle.

%\section{Hyperparameters}\label{sec:appendix_hyperparameters}

\end{document}